\documentclass[table]{gtech}
\PassOptionsToPackage{table, usenames, dvipsnames}{xcolor}

\usepackage{xcolor}
\usepackage{soul}
\usepackage{amssymb}
\usepackage{multirow}
\usepackage{bigdelim}
\usepackage{longtable}
\usepackage{tabularray}
\usepackage{wrapfig}
\usepackage[most]{tcolorbox}
\usepackage{url}
\usepackage{float}
\usepackage{datatool}
\usepackage{enumitem}
\usepackage{subcaption} % 支持子图标题
\usepackage[justification=centering]{caption}

\RequirePackage{tgpagella} % text only
\RequirePackage{mathpazo}  % math & text
\RequirePackage{inconsolata} % for tt font
\usepackage{makecell}

\usepackage{adjustbox}
\usepackage{tablefootnote}
\usepackage{threeparttable}

\usepackage{booktabs} % 用于 \toprule, \midrule, \bottomrule
\usepackage{array}    % 用于定义新的列类型
\newcolumntype{C}[1]{>{\centering\arraybackslash}m{#1}}

\usepackage[utf8]{inputenc} % allow utf-8 input
\usepackage[T1]{fontenc}    % use 8-bit T1 fonts
\usepackage{hyperref}       % hyperlinks
\usepackage{url}            % simple URL typesetting
\usepackage{booktabs}       % professional-quality tables
\usepackage{amsfonts}       % blackboard math symbols
\usepackage{nicefrac}       % compact symbols for 1/2, etc.
\usepackage{microtype}      % microtypography
\usepackage{xcolor}         % colors
\usepackage{xspace}
\usepackage{amsthm}   % 必须用于 definition 环境
\usepackage{amsmath}  % 高级数学公式
\usepackage{amssymb}  % 数学符号扩展
\usepackage{bm}       % 粗体数学符号（可选）

\theoremstyle{definition}

\theoremstyle{plain}

\renewcommand{\title}[1]{\newcommand{\titlelist}{{\LARGE\bfseries #1}}}

\newcommand{\ignore}[1]{}

\usepackage{arydshln}
\definecolor{CQColor}{rgb}{0.0,0.0,1.0} % color for Aaron

\usepackage{graphicx}
\usepackage{colortbl}
\usepackage{amssymb}
\usepackage{pifont}
\usepackage{booktabs,multirow}
\usepackage{makecell}
\usepackage{tabulary}
\usepackage{fontawesome5}
\usepackage{bbding}
\usepackage{multicol}

\newlength\savewidth

\newcommand{\heroname}{UFP4\xspace}              % final hero recipe name; 写作统一引用此 macro
\newcommand{\tetrajet}{TetraJet-v2\xspace}        % TetraJet-v2

\newcommand{\fwdy}{\texttt{fwd\_y}\xspace}
\newcommand{\bwddx}{\texttt{bwd\_dx}\xspace}
\newcommand{\bwddw}{\texttt{bwd\_dw}\xspace}

\title{Rethinking Shrinkage Bias in LLM FP4 Pretraining:\\Geometric Origin, Systemic Impact, and UFP4 Recipe}

\author[]{\vspace{0.5em}Qian Zhao}
\author[]{Kunlong Chen}
\author[]{Changxin Tian}
\author[]{Zhonghui Jiang}
\author[]{Haitao Zhang}
\author[]{Chaofan Yu}
\author[]{Peijie Jiang}
\author[]{Mingliang Gong}
\author[]{Jia Liu}
\author[]{Ziqi Liu}
\author[*]{Zhiqiang Zhang}
\author[]{Jun Zhou}

\affiliation[]{Ling Team, Ant Group}
\contribution[*]{Corresponding author}
\abstract{
FP4 training promises substantial reductions in memory and computation cost for LLM pretraining, yet current FP4 hardware paths and recipes, including NVIDIA Blackwell/Rubin-class systems and AMD MI350-series GPUs, remain centered on E2M1 data elements.
In this study, we identify a fundamental limitation of that choice: non-uniform formats such as E2M1 inherently suffer from \emph{Shrinkage Bias}, a systematic negative rounding error caused by the geometric asymmetry of their representable bins.
We show that this bias accumulates multiplicatively across layers and is amplified by the Random Hadamard Transform (RHT), providing a unified explanation for the training instability observed in existing E2M1-based FP4 recipes.
In contrast, uniform grids (E1M2/INT4) bypass this grid-geometry error and better convert the improved bucket utilization from RHT into higher quantization quality.
Based on this finding, we propose \heroname{}, a uniform 4-bit training recipe that applies RHT to all three training GEMMs while restricting stochastic rounding to \(dY\) alone.
On Dense 1.5B, MoE 7.9B, and MoE 124B long-run pretraining, \heroname{} consistently achieves lower BF16-relative loss degradation than strong E2M1-based baselines, supported by scaling-law analysis and ablation studies.
Our results suggest that future accelerators should support E1M2/INT4-style uniform 4-bit grids as first-class training primitives alongside E2M1.
}

\date{\today\vspace{-1mm}}
\gtechdata[Correspondence]{\email{\{zq317110,lingyao.zzq\}@antgroup.com}}

\begin{document}
\maketitle

\begin{figure}[b]
    \centering
    \hfill
    \begin{subfigure}[b]{0.7\textwidth}
        \includegraphics[width=1.\linewidth]{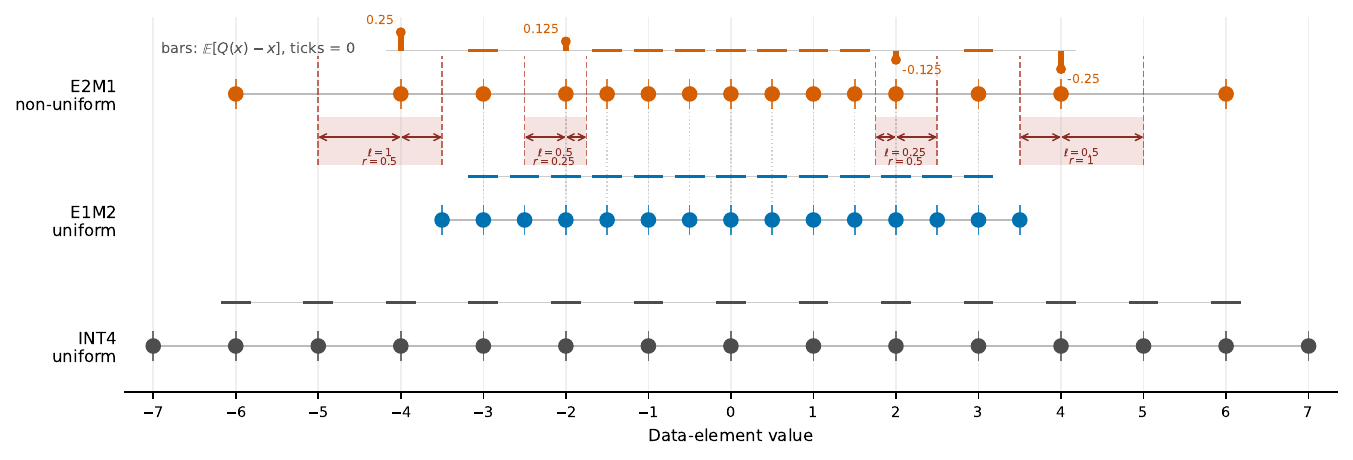}
        \caption{Representable values and RTNE bins of E2M1 vs.\ E1M2/INT4.}
        \label{fig:fp4-int4-format-codebooks}
    \end{subfigure}
    \hfill
    \begin{subfigure}[b]{0.26\textwidth}
        \includegraphics[width=\textwidth]{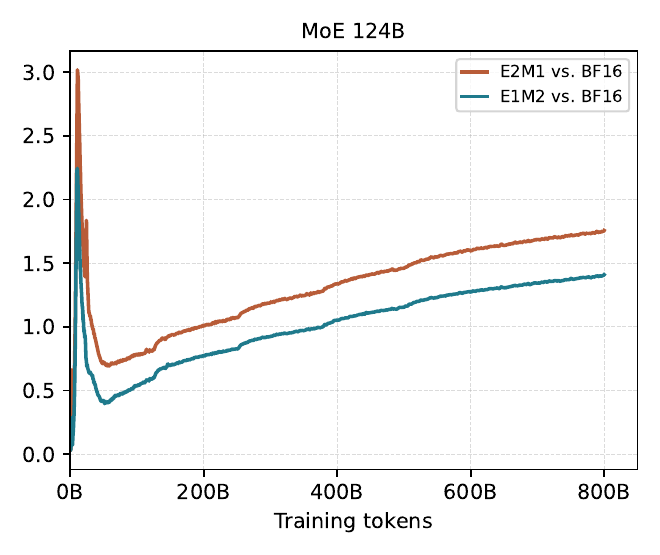}
        \caption{BF16-relative loss degradation (124B MoE).}
        \label{fig:intro-loss-diff}
    \end{subfigure}
    \hfill
    \caption{(a) Red spans mark RTNE rounding bins; bars show expected rounding error per bin. E2M1's non-uniform bins exhibit systematic toward-zero bias (Shrinkage Bias) due to geometric asymmetry, while E1M2/INT4 uniform bins remain unbiased. (b) This grid-level advantage translates to training quality: on 124B MoE long-run pretraining, E1M2-based \heroname{} significantly outperforms E2M1-based baselines in BF16-relative loss degradation.}
\end{figure}

\section{Introduction}
\label{sec:intro}

% Scaling Law 的提出与应用，使得小规模模型上的架构、数据和训练策略创新能够更可靠地外推到大规模训练，从而推动了大语言模型能力上界的持续提升。然而，完整训练一个前沿模型的成本依然极高，显存、带宽和矩阵计算吞吐已经成为继续 scaling 的关键瓶颈。因此，如何降低训练过程中的资源和显存需求，已经成为推动 LLM 继续扩展的重要问题。
% 低精度训练是降低训练成本的一条重要路径。其基本思想是在前向传播和反向传播过程中使用低精度 activation、weight 和 gradient，并借助现代硬件高效的低精度矩阵计算能力提升端到端训练吞吐。近年来，FP8 已经在多个大规模训练系统中被采用 \citep{dpsk,ling}，表明进一步降低训练精度具有现实可行性。随着硬件发展，Blackwell 等新一代加速器开始提供原生 FP4 矩阵计算路径，为进一步降低 LLM 训练成本提供了新的机会。
% 当前主流 FP4 训练接口大多围绕 E2M1 format 设计。例如，MXFP4 / NVFP4 类格式会为细粒度 tensor block 维护独立 scale，并通过不同 block size 和 scale hierarchy 在动态范围、精度和硬件效率之间折中。NVFP4 由于更小的 block 和更精细的 scale 设计，通常比 MXFP4 具有更好的训练精度。即便如此，FP4 只有 16 个可表示值，直接用于训练仍会面临严重的收敛和稳定性问题。为缓解这些问题，NVFP4 recipe 引入 Random Hadamard Transform（RHT）和 stochastic rounding 来处理 outlier 与梯度更新不稳定；后续工作则进一步从 scale 设计、rounding / gradient estimation、outlier control 和量化感知训练等方向缩小 FP4 与 BF16 之间的差距。
% 尽管现有方法显著改善了 E2M1-based FP4 training，它们大多仍默认保留 E2M1 data-element grid。本文重新审视这一默认选择。我们的核心发现是：RHT 不仅缓解 outlier，也改变了量化问题本身。RHT 将 tensor 从 dynamic-range-limited regime 推向 local-resolution-limited regime；在这一 regime 下，E2M1 的非均匀 RTNE bin 会产生 Shrinkage Bias，使 RHT 带来的 bucket utilization 提升在关键 tensor 上难以转化为更低的量化误差。相比之下，E1M2/INT4-style uniform grid 能够更好地利用 post-RHT 分布，从而获得更高的量化质量。
% 基于这一发现，我们提出 UFP4，一种面向训练的 uniform 4-bit recipe。UFP4 使用 E1M2/INT4-style uniform grid，在 fwd_y、bwd_dx 和 bwd_dw 三个 training GEMM 上全部启用 RHT，并仅在 dY 上使用 stochastic rounding。UFP4 表明，full-RHT training 本身并非问题；问题在于 E2M1 grid 与 post-RHT tensor regime 的不匹配。在 uniform grid 下，RHT 可以扩展到 forward、data-gradient 和 weight-gradient 路径，并在长跑训练中保持稳定收益。

Continued scaling of Large Language Models (LLMs) has pushed the frontier of model capabilities, yet the rapidly growing cost, memory footprint, and energy consumption of pretraining demand more efficient numerical formats~\citep{hoffmann2022trainingcomputeoptimal,zhao2026survey}.
Low-precision training has therefore become a key direction for reducing training cost, with FP8 already adopted by several large-scale training systems~\citep{DeepSeekAI2024DeepSeekV3TR,Team2025EveryAB}.
The basic idea is to represent activations, weights, and gradients in lower precision during both forward and backward passes, leveraging the high-throughput low-precision matrix-multiply units on modern accelerators~\citep{micikevicius2022fp8formatsdeeplearning}.
More recently, accelerators such as NVIDIA Blackwell expose native FP4 computation paths~\citep{nvidia2026pretraininglargelanguagemodels}, offering a new opportunity to further halve the precision and reduce training cost.
% LLM 在模型规模和训练数据规模上的持续扩展推动了其能力的前沿。然而，预训练过程中快速增长的计算成本、内存占用和能耗，也对更加高效的数值格式提出了迫切需求。因此，以 FP8 为代表的低精度训练已成为降低训练成本的重要方向，被多个大规模训练系统采用。其基本思想是在前向传播和反向传播过程中，以低精度格式表示 activation、weight 和 gradient，并利用现代加速器上高吞吐的低精度矩阵计算单元。随着 Blackwell 等新一代加速器开始提供原生 FP4 矩阵计算路径，FP4 训练为进一步降低 LLM 预训练成本提供了新的机会。

Current FP4 training interfaces are largely designed around E2M1, a 4-bit floating-point format with 2 exponent bits and 1 mantissa bit.
For example, MXFP4~\citep{rouhani2023microscalingdataformatsdeep} and NVFP4 \citep{nvidia2026pretraininglargelanguagemodels} maintain independent scale factors for fine-grained tensor blocks, trading off dynamic range, precision, and hardware efficiency through different block sizes and scale hierarchies.
Compared with MXFP4, NVFP4 typically provides better training accuracy through smaller blocks and a more refined scaling design.
Nevertheless, with only 16 representable values, end-to-end FP4 training remains challenging: without additional stabilization, FP4 recipes suffer from severe convergence issues and loss degradation relative to BF16~\citep{2025arXiv250117116W,chmiel2025fp4wayfullyquantized,nvidia2026pretraininglargelanguagemodels}.
To mitigate these problems, the NVFP4 recipe introduces the Random Hadamard Transform (RHT) and stochastic rounding (SR) to handle outliers and stabilize gradient updates~\citep{3737916.3741096,pmlr-v37-gupta15}.
Subsequent work further reduces the gap between FP4 and BF16 training through improved scale design, rounding or gradient-estimation methods, outlier control, and quantization-aware training~\citep{chen2026tetrajetv2accuratenvfp4training,panferov2026quartetiiaccuratellm,cao2025metis,cook2026sixaccuratenvfp4quantization,li2026faarformatawareadaptiverounding}.
% 当前主流 FP4 训练接口大多围绕 E2M1 format 设计。例如，MXFP4 / NVFP4 类格式会为细粒度 tensor block 维护独立 scale，并通过不同 block size 和 scale hierarchy 在动态范围、精度和硬件效率之间折中。NVFP4 由于更小的 block 和更精细的 scale 设计，通常比 MXFP4 具有更好的训练精度。即便如此，FP4 只有 16 个可表示值，直接用于训练仍会面临严重的收敛和稳定性问题。为缓解这些问题，NVFP4 recipe 引入 Random Hadamard Transform（RHT）和 stochastic rounding 来处理 outlier 与梯度更新不稳定；后续工作则进一步从 scale 设计、rounding / gradient estimation、outlier control 和量化感知训练等方向缩小 FP4 与 BF16 之间的差距。

Despite effectiveness, most existing methods still retain the E2M1 data-element format unchanged.
This paper revisits that default choice. Our central observations are:
\textbf{(1)}~Non-uniform formats such as E2M1 inherently suffer from \emph{Shrinkage Bias}—a negative expected rounding error under Round-to-Nearest-Even (RTNE) caused by the geometric asymmetry of their representable bins.
\textbf{(2)}~This bias accumulates multiplicatively across layers, causing systematic signal decay. Moreover, RHT exacerbates the problem under E2M1 by shifting tensor mass into the most asymmetric bins, further degrading training stability.
In contrast, as illustrated in Figure~\ref{fig:fp4-int4-format-codebooks}, uniform grids (e.g., E1M2/INT4) bypass this grid-geometry error entirely, better match the post-RHT distribution, and convert improved bucket utilization into consistently higher quantization quality.
% 尽管有效，现有方法大多仍保留 E2M1 data-element 格式不变。本文重新审视这一默认选择，核心发现如下：(1) E2M1 等非均匀格式固有地存在 Shrinkage Bias——由其表示 bin 的几何不对称导致的 RTNE 负期望舍入误差。(2) 该偏差在层间乘性累积，造成系统性信号衰减；而 RHT 会将 tensor mass 推入最不对称的 bin，在 E2M1 下进一步恶化训练稳定性。相比之下，uniform grid（如 E1M2/INT4）完全规避了这一 grid-geometry 误差，更好地匹配 post-RHT 分布，将 bucket utilization 的提升转化为一致更高的量化质量。

Based on this finding, we propose \heroname{}, a uniform 4-bit training recipe built on an E1M2/INT4-style grid.
\heroname{} applies RHT to the operands of all three linear-layer training GEMMs, i.e., forward (FPROP, \fwdy{}), data-gradient (DGRAD, \bwddx{}), and weight-gradient (WGRAD, \bwddw{}), and uses stochastic rounding only when quantizing the upstream gradient \(dY\).
The recipe demonstrates that full-RHT training is not inherently harmful; rather, the issue lies in the mismatch between E2M1 and the post-RHT tensor regime.
With a uniform grid, RHT extends to all three paths while preserving stable long-run gains: on a 124B MoE long-run pretrain, E1M2-based \heroname{} significantly outperforms strong E2M1-based baselines in BF16-relative loss degradation (Figure~\ref{fig:intro-loss-diff}).
% 基于这一发现，我们提出 UFP4，一种基于 E1M2/INT4-style uniform grid 的 4-bit 训练 recipe。UFP4 在 forward、data-gradient 和 weight-gradient 三个 training GEMM 上全部启用 RHT，并仅在 dY 上使用 stochastic rounding。该 recipe 表明，full-RHT training 本身并非问题；问题在于 E2M1 与 post-RHT tensor regime 的不匹配。在 uniform grid 下，RHT 可扩展到全部三条路径并保持稳定的长跑收益：在 124B MoE 长跑预训练中，E1M2-based UFP4 在 BF16-relative loss degradation 上显著优于 E2M1-based 强 baseline（Figure~\ref{fig:intro-loss-diff}）。

We summarize our contributions as follows:
\begin{itemize}
    \item We identify and formalize \emph{Shrinkage Bias}: a systematic negative rounding error inherent to non-uniform grids such as E2M1, caused by the geometric asymmetry of their RTNE bins.

    \item We theoretically and empirically show that Shrinkage Bias accumulates multiplicatively across layers and is amplified by RHT, providing a unified explanation for the training instability observed in existing E2M1-based FP4 recipes, e.g., NVFP4 recipe.

    \item We propose \heroname{}, a 4-bit training recipe built on uniform E1M2/INT4-style grids that enables RHT on all three training GEMMs while restricting stochastic rounding to \(dY\) alone.

    \item We validate \heroname{} on Dense 1.5B, MoE 7.9B, and MoE 124B long-run pretraining, supported by scaling-law analysis, ablation studies, and fused-kernel benchmarks, demonstrating that uniform grids are practical first-class FP4 training formats at industrial scale.
\end{itemize}
\section{Preliminaries}
\label{sec:preliminaries}
% 中文意图：本章只建立记号和背景，不提前下结论。需要让读者理解 E2M1/E1M2/INT4、blockwise quantization、RHT 以及训练中三个 GEMM 的形式。避免在这里写 “E2M1 错了” 或 “UFP4 最优”。

\subsection{4-bit Formats and Blockwise Quantization}
\label{sec:prelim-quant}
% 中文意图：介绍 E2M1、E1M2、INT4 的 codebook 差异，并定义统一的 blockwise quantizer。核心是把后文所有 format comparison 固定到同一 scale convention 下。

FP4 formats use one sign bit and split the remaining bits into exponent and mantissa fields, denoted E$x$M$y$.
We consider two FP4 formats, E2M1 and E1M2, and an INT4 codebook (\Cref{fig:fp4-int4-format-codebooks}).

In practice, blockwise quantization is widely adopted to improve precision by partitioning a tensor \(\mathbf T\) into contiguous blocks \(\{B\}\) and mapping each element to a codebook level under a shared per-block scale.
Let \(G=\{g\}\) denote the normalized codebook of a chosen format, \(g_{\max}=\max_{g\in G}|g|\) its largest magnitude level, and \(\rho_G\) the rounding rule.  The quantization process is then:
\[
    s_B=\frac{\max_{x_j\in B}|x_j|}{g_{\max}},
    \qquad
    q_i=\rho_G\!\left(\frac{x_i}{s_B}\right)\in G.
\]
We write \(Q_G(\mathbf T)\) for the quantized tensor; in arithmetic expressions it denotes the dequantized numerical tensor.\footnote{%
  MXFP4 uses \(1{\times}32\) blocks with an E8M0 scale; NVFP4 uses \(1{\times}16\) blocks with two-level scaling.}
The rounding rule \(\rho_G\) is either Round-To-Nearest-Even (RTNE) or Stochastic Rounding (SR).
RTNE selects the nearest level in \(G\) with deterministic tie-breaking.
When \(x_i/s_B\) falls between adjacent levels \(g_a<g_b\), SR samples:
\[
    \Pr[q_i=g_b]=\frac{x_i/s_B - g_a}{g_b-g_a},
    \qquad
    \Pr[q_i=g_a]=\frac{g_b - x_i/s_B}{g_b-g_a}.
\]
SR preserves the normalized value in expectation, while RTNE is deterministic. 
In practice, RTNE is widely adopted by advanced low-precision methods such as NVFP4 Recipe~\citep{nvidia2026pretraininglargelanguagemodels}.

\subsection{Random Hadamard Transforms}
\label{sec:prelim-rht}
% 中文意图：介绍 RHT 是 norm-preserving 的 rotation，能把 outlier energy 分散到更多坐标中。随后给出 fwd_y、bwd_dx、bwd_dw 三个 GEMM 加 RHT 后的公式，为 Shrinkage Bias 和 UFP4 recipe 做铺垫。

Real training tensors often contain outlier coordinates, causing most codebook levels to be underutilized. Random Hadamard Transforms (RHT) address this by applying a norm-preserving rotation that disperses outlier energy across all coordinates before quantization.

Concretely, we use the Sylvester Hadamard matrices defined recursively as
\[
    \mathbf H_{2n}=\frac{1}{\sqrt{2}}
    \begin{bmatrix}
        \mathbf H_n & \mathbf H_n\\
        \mathbf H_n & -\mathbf H_n
    \end{bmatrix},
    \qquad \mathbf H_1=[1],
\]
where \(\mathbf H_n^\top\mathbf H_n=I_n\) for \(n=2^k\).
An RHT additionally applies a random sign matrix \(\mathbf S_n=\operatorname{diag}(\epsilon_1,\ldots,\epsilon_n)\), \(\epsilon_i\in\{-1,+1\}\), before the Hadamard transform.
Since \(\mathbf H'_n=\mathbf S_n\mathbf H_n\) is orthogonal, applying the same transform to the shared GEMM dimension preserves the full-precision result:
\begin{gather*}
    \mathbf H'_n=\mathbf S_n\mathbf H_n, \\
    \mathbf Y=\mathbf X\mathbf W^\top
    =
    (\mathbf X\mathbf H'_n)(\mathbf W\mathbf H'_n)^\top .
\end{gather*}
The corresponding low-precision GEMM quantizes the rotated operands:
\[
    \widehat{\mathbf Y}
    =
    Q_G(\mathbf X\mathbf H'_n)\,
    Q_G(\mathbf W\mathbf H'_n)^\top .
\]
RHT reduces outlier dominance and improves codebook utilization; however, its net effect on quantization error depends on the FP4 grid, which we analyze in \Cref{sec:shrinkage-after-rht}.
For the three training GEMMs, RHT is applied along the shared reduction dimension: the input-channel dimension for \fwdy{}, the output-channel dimension for \bwddx{}, and the batch-token dimension for \bwddw{}.

% ======================================================================
\section{Shrinkage Bias in 4-bit Format Grids}
\label{sec:shrinkage-bias}
In this section, we investigate the systematic \emph{Shrinkage Bias} inherent to non-uniform quantization formats like E2M1, driven by the geometric asymmetry of their rounding bins. We demonstrate how this bias accumulates multiplicatively across deep networks and expose a hidden pitfall: standard outlier-mitigation techniques like the RHT inadvertently exacerbate this instability when paired with non-uniform grids.

\subsection{Geometric Origin: Shrinkage Bias from Asymmetric RTNE Bins}
\label{sec:cell-contraction}
% 中文意图：用 rounding-bin asymmetry 形式化 E2M1 的 Shrinkage Bias。重点不是“任意分布都会 shrink”，而是 E2M1 的某些 codepoint 不在自己的 RTNE bin 中心；因此即使 bin 内局部分布均匀，也会产生非零期望幅值误差。E1M2/INT4 的均匀 interior bins 消除了这个 grid-geometry source。该节只讨论固定 scale、无 clipping 的 RTNE 机制；SR、clipping、scale estimation 作为额外因素留给实验和讨论。

Consider a block with scale $s>0$, and let $t=|x|/s$ denote the normalized magnitude. 
We define \emph{Shrinkage Bias} as a negative expected Round-to-Nearest-Even (RTNE) error in the normalized magnitude space. 
Specifically, for a distribution $\mathcal{P}$ over $t$, let the expected error $b_G(\mathcal{P})=\mathbb{E}_{t\sim\mathcal{P}}[\rho_G(t)-t]$, where $\rho_G(\cdot)$ is the RTNE mapping to the codebook $G$. A value $b_G(\mathcal{P})<0$ indicates Shrinkage Bias.

Assuming no clipping ($t \le \max(G)$), we focus on the non-negative normalized levels of the format:
$$
    G_+ = \{q_0,q_1,\ldots,q_n\},
    \qquad
    0 = q_0 < q_1 < \cdots < q_n.
$$
For an interior quantization level $q_i$ ($i \in \{1, \dots, n-1\}$), the interior of its RTNE rounding bin is
\begin{equation}
    \mathcal{B}_i
    =
    \left(
    \frac{q_{i-1}+q_i}{2},
    \frac{q_i+q_{i+1}}{2}
    \right),
    \qquad
    \ell_i=\frac{q_i-q_{i-1}}{2},
    \quad
    r_i=\frac{q_{i+1}-q_i}{2},
    \label{eq:rtn-bin}
\end{equation}
where $\ell_i$ and $r_i$ are the left and right bin widths around $q_i$. 
For $t\in\mathcal{B}_i$, the magnitude error is $\rho_G(t)-t=q_i-t$.
If the density inside this bin is locally uniform, we can calculate the conditional expected error by changing variables to $u = t - q_i$:
\begin{equation}
    \mathbb{E}\!\left[\rho_G(t)-t \mid t\in\mathcal{B}_i\right]
    =
    \frac{1}{\ell_i+r_i}
    \int_{-\ell_i}^{r_i} (-u)\,du
    =
    \frac{\ell_i-r_i}{2}
    =
    \frac{2q_i-q_{i-1}-q_{i+1}}{4}.
    \label{eq:bin-asymmetry-bias}
\end{equation}
Consequently, an asymmetric RTNE bin with $r_i > \ell_i$ inherently yields a negative expected error under uniform local density. This reveals a grid-geometry source of magnitude shrinkage, distinct from distribution-induced quantization error inside a symmetric bin.

As the mainstream low-precision format utilized in advanced pipelines (e.g., NVFP4), E2M1 inherently contains such asymmetric bins at spacing-transition points.
For example, given the non-negative E2M1 magnitudes $\{0, 0.5, 1, 1.5, 2, 3, 4, 6\}$, the level $q_i=2$ yields a conditional bias of $-0.125$, and $q_i=4$ exhibits the same asymmetry at double the scale.\footnote{For $q_i=2$: bin $(1.75,2.5)$, $\ell_i=0.25$, $r_i=0.5$; Eq.~\eqref{eq:bin-asymmetry-bias} gives $\mathbb{E}[\rho_G(t)-t \mid t\in(1.75,2.5)]=-0.125$.}
By contrast, uniform grids (e.g., E1M2) satisfy $\ell_i=r_i$ for all bins, thereby eliminating this fundamental source of bias.
We visually corroborate this contrast in \Cref{fig:fp4-int4-format-codebooks}, which compares the E2M1, E1M2, and INT4 grids to explicitly illustrate both the geometric origin of Shrinkage Bias and its resulting error distribution.

\begin{tcolorbox}[colback=gray!5,colframe=gray!50,arc=0pt,outer arc=0pt,leftrule=3pt,rightrule=0pt,toprule=0pt,bottomrule=0pt,top=2mm,bottom=2mm]
\textbf{Key Finding 1:} Non-uniform quantization formats (e.g., E2M1) inherently suffer from \emph{Shrinkage Bias} due to the geometric asymmetry of their rounding bins, whereas uniform grids (e.g., E1M2) bypass this fundamental source of grid-geometry error.
\end{tcolorbox}

\subsection{Systemic Impact: Propagation and RHT Exacerbation}

This bin-level bias is not just harmless zero-mean noise. When it enters matrix multiplications (GEMMs), it acts as a systematic attenuation that cascades through the network.

% \subsection{Propagation through Matrix Multiplication}
\paragraph*{Propagation through Multiplicative Accumulation.}
\label{sec:gemm-bias-propagation}
% 中文意图：用严格的 projection decomposition 表达 operand-level attenuation 如何进入 GEMM 主信号。单次 GEMM 部分是恒等式；多次 GEMM 部分只作为 residual-incoherence 假设下的 propagation model，不写成无条件训练收敛定理。

The bin-level bias discussed above becomes critical because it survives tensor aggregation and accumulates across deep paths. 
For a GEMM $Z=AB^\top$ with quantized operands $\widehat{A}=Q_G(A)$ and $\widehat{B}=Q_G(B)$, we measure magnitude shrinkage by projecting the quantized operands onto their exact BF16 counterparts:
\begin{equation}
    \alpha_A
    =
    \frac{\langle \widehat{A},A\rangle_F}{\|A\|_F^2},
    \qquad
    \widehat{A}
    =
    \alpha_A A + R_A,
    \qquad
    \langle R_A,A\rangle_F=0,
    \label{eq:operand-projection-a}
\end{equation}
and analogously $\widehat{B}=\alpha_B B+R_B$. A scaling factor $\alpha_A<1$ indicates a reduced signal component aligned with $A$, acting as the tensor-level manifestation of magnitude shrinkage.

Substituting these orthogonal decompositions into the GEMM yields the exact identity:
\begin{equation}
    Z_q
    =
    \widehat{A}\widehat{B}^{\top}
    =
    \underbrace{\alpha_A\alpha_B AB^\top}_{\alpha_A\alpha_B Z}
    +
    \underbrace{\alpha_A A R_B^\top
    +
    \alpha_B R_A B^\top
    +
    R_A R_B^\top}_{\text{residual noise}}.
    \label{eq:gemm-propagation}
\end{equation}
When the residual terms are incoherent with $Z$, the principal signal is attenuated by a factor of $\eta \approx \alpha_A\alpha_B < 1$. 
This decay propagates systematically. For a sequence of $K$ quantized GEMMs, let $\eta_k \approx \alpha_{A,k}\alpha_{B,k}$ denote the coherent attenuation factor of the $k$-th operation. The initial clean signal is cumulatively scaled by:
\begin{equation}
    \prod_{k=1}^{K}\eta_k
    =
    \prod_{k=1}^{K}(1-\delta_k)
    \approx
    \exp\!\left(-\sum_{k=1}^{K}\delta_k\right),
    \qquad
    \delta_k=1-\eta_k .
    \label{eq:multi-gemm-attenuation}
\end{equation}
Thus, unlike zero-mean noise which tends to cancel out, this systematic shrinkage accumulates multiplicatively (see \Cref{sec:appendix-propagation} for the full derivation). Crucially, while quantization errors in \bwddw{} are leaf-gradients directly consumed by the optimizer, errors in non-leaf outputs (\fwdy{} and \bwddx{}) cascade through subsequent layers, actively compounding this systematic decay.

% \subsection{Shrinkage Bias after RHT}
\paragraph*{Exacerbation under RHT.}
\label{sec:shrinkage-after-rht}
% 中文意图：解释 RHT 后为什么问题暴露。pre-RHT 下 E2M1 dynamic range 合理；post-RHT 后 tensor 变 concentrated，local resolution 主导，E2M1 的宽 cell 变成问题。加入 shrinkage-region mass 和 bucket entropy 两个指标，服务后续真实 tensor 分析。
\begin{figure}[t]
    \centering
    \includegraphics[width=\linewidth]{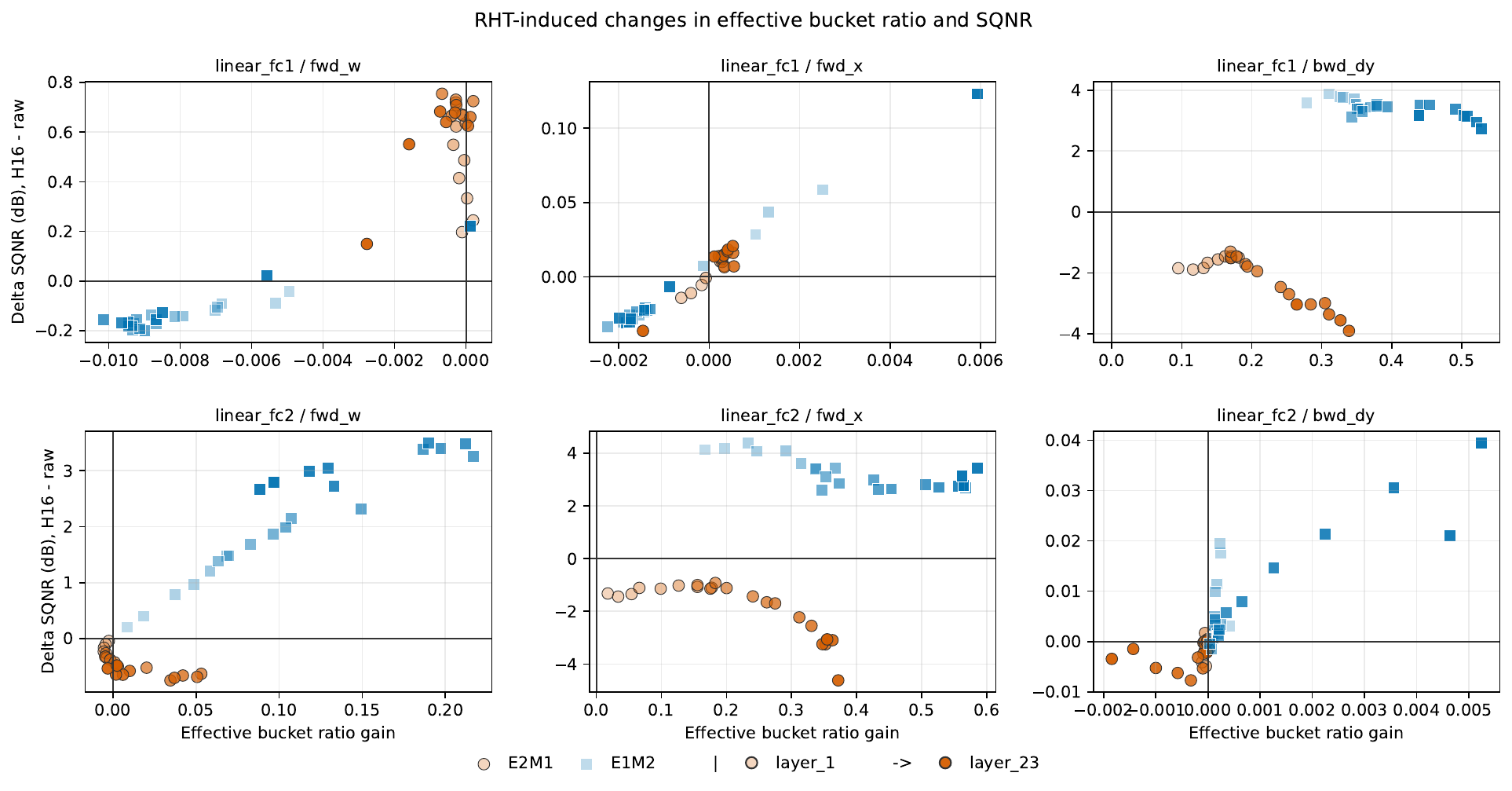}
    \caption{
    {Impact of RHT on effective bucket ratio and SQNR.} 
    Colors distinguish the quantization formats (E2M1 vs. E1M2), and marker opacity increases with network depth. Evaluated on real linear-layer tensors: weights (\texttt{fwd\_w}), inputs (\texttt{fwd\_x}), and output gradients (\texttt{bwd\_dy}).
    }
    \label{fig:rht-bucket-sqnr-tradeoff}
\end{figure}

While E2M1's wide dynamic range is initially a sensible choice for mitigating saturation in outlier-heavy tensors, RHT fundamentally alters this regime. By spreading outlier energy across coordinates, RHT transforms the tensor from being dynamic-range-limited to local-resolution-limited. Consequently, the bottleneck shifts from representing extreme outliers to accurately preserving the dense probability mass at typical magnitudes.

To formalize this shift, we measure utilization across $K$ quantized magnitude buckets. Given empirical fractions $p_i$ for each bucket (including zero), we define the bucket entropy $\mathcal{E}(G,T)$ and report the effective bucket ratio $B_{\mathrm{eff}}(G,T)$:
\begin{equation}
    B_{\mathrm{eff}}(G,T)
    =
    \frac{\exp(\mathcal{E}(G,T))}{K},\;\;\;\;\ \operatorname{where} \;\;\;\; \mathcal{E}(G,T) = -\sum_{i=1}^{K} p_i \log(p_i+\epsilon)
    \label{eq:effective-bucket-ratio}
\end{equation}
where $B_{\mathrm{eff}} \in [1/K, 1]$ spans from single-bucket collapse to uniform utilization. 
To assess quantization quality, we compute the Normalized MSE, $\mathrm{NMSE}_{A}(G,T) = \|Q_G(TA)-TA\|_F^2 / \|TA\|_F^2$. Setting $A=I$ and $A=\mathbf{H}_{16}$ yields the raw and post-RHT errors, respectively, allowing us to summarize the RHT-induced change via the Signal-to-Quantization-Noise Ratio:
\begin{equation}
    \Delta\mathrm{SQNR}
    =
    10\log_{10}
    \frac{\mathrm{NMSE}_{I}}{\mathrm{NMSE}_{\mathbf{H}_{16}}},
    \label{eq:delta-sqnr}
\end{equation}
where $\Delta\mathrm{SQNR} > 0$ indicates that rotation improves tensor-level fidelity.

\Cref{fig:rht-bucket-sqnr-tradeoff} analyzes real training tensors from MLP blocks (\texttt{fwd\_w}, \texttt{fwd\_x}, and \texttt{bwd\_dy}). 
As expected, RHT substantially increases $B_{\mathrm{eff}}$ for outlier-heavy tensors (e.g., \texttt{linear\_fc1/bwd\_dy} and \texttt{linear\_fc2/fwd\_x}), consistent with the outlier-amplifying behavior of SwiGLU \citep{Jiang2026PowLUAA}. 
Crucially, however, this improved utilization yields divergent outcomes depending on grid geometry. By shifting tensor mass from extreme tails into mid-magnitude regions, RHT inadvertently forces data into E2M1's most asymmetric rounding bins, exacerbating quantization noise ($\Delta\mathrm{SQNR} < 0$). In contrast, the uniformly spaced E1M2 grid safely translates this flattened distribution into higher fidelity ($\Delta\mathrm{SQNR} > 0$). 
This exposes a hidden pitfall: despite effectively dispersing outliers, applying RHT to non-leaf paths (\texttt{fwd\_y}, \texttt{bwd\_dx}) systematically degrades E2M1-based recipes.

\begin{tcolorbox}[colback=gray!5,colframe=gray!50,arc=0pt,outer arc=0pt,leftrule=3pt,rightrule=0pt,toprule=0pt,bottomrule=0pt,top=2mm,bottom=2mm]
\textbf{Key Finding 2:} Shrinkage Bias accumulates multiplicatively across network layers, causing systematic signal decay. However, applying RHT exacerbates this decay under E2M1 by shifting tensor mass into the most biased asymmetric bins, degrading deep training stability.
\end{tcolorbox}

% \subsection{Community remedies and their scope}
% \label{sec:community-remedies}
% 中文意图：这里不写长 Related Work，只用 1 段把社区方法按层级分类。重点是说明很多工作缓解的是 rounding、scale、tensor distribution，但保留 E2M1 grid，因此不能从根上消除本文定义的 grid-induced Shrinkage Bias。所有具体论文引用后续由 Related Work 补。

% Existing FP4 training methods can be viewed as attempts to mitigate the same failure mode at different levels.  Quantizer-side methods modify rounding rules, scale updates, or backward alignment to reduce deterministic error or variance.  Tensor-side methods decompose, smooth, or separate difficult components before quantization.  These techniques can be valuable and are largely orthogonal to \heroname{}.  However, if the final data-element grid remains E2M1, the residual tensor is still mapped onto a non-uniform grid whose geometry can produce post-RHT contraction.  \heroname{} therefore takes the complementary route: it changes the data-element grid itself.

\begin{figure}[t]
    \centering
    \includegraphics[width=\linewidth]{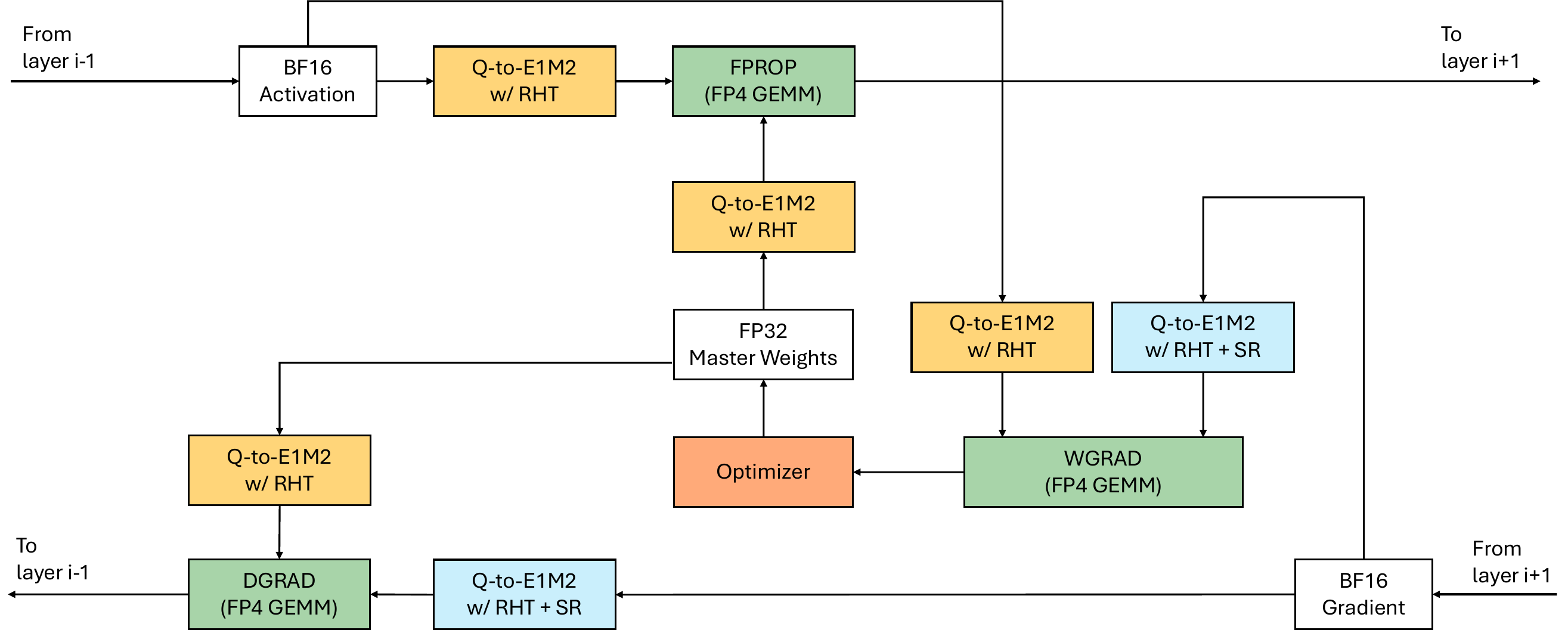}
    \caption{
    {Overview of the \heroname{} recipe.} By leveraging a uniform grid, \heroname{} enables pervasive RHT across all forward and backward GEMMs.
    }
    \label{fig:ufp4-recipe-overview}
\end{figure}

% ======================================================================
\section{\heroname{}: 4-bit Training Recipe with Uniform Grids}
\label{sec:ufp4}

% 当下广泛被采用的 FP4 Recipe 大都基于 E2M1 的数据格式，受限于 IEEE 754 标准及软硬件约束，社区中诸多缓解 FP4 训练误差的方法均无法从根源上消除 Shrinkage Bias。
% UFP4 是一种建立在 E1M2/INT4 数据格式基础上的 4-bit 训练 Recipe，其核心思路如图所示：
% 1. 对于 fwd_x, fwd_w, bwd_dy 等 tensor 在 GEMM 计算前均执行 RHT 变换并量化到 INT4 精度中
% 2. 此外，SR 应用于 bwd_dy 来保障梯度更新方向的无偏性
% UFP4 Recipe 在 RHT + 等间距 E1M2/INT4 格式下提供了一种新的 4bit 训练 Recipe 的视角，它足够的简单、高效且具备更优的训练精度。我们倡导各家厂商重新重视 E1M2/INT4，并在未来的高性能训练卡上完成支持。
% 针对 scale 精度选择与本文关注的目标是正交的（例如 scale 使用一级量化（mxfp4 1*16 E8M0 scale）或者二级量化（nvfp4 1*16 E4M3 scale + global FP32 scale）或者三级量化（HiFloat4）），scale 的设计应该在精度可靠范围内选择最贴合硬件计算效率的方案。

Most existing FP4 training recipes default to the E2M1 format \citep{nvidia2026pretraininglargelanguagemodels,chen2026tetrajetv2accuratenvfp4training,panferov2026quartetiiaccuratellm}. 
While initially attractive for raw outlier-heavy tensors, current mitigation techniques (e.g., specialized rounding, scaling, or tensor preprocessing) merely treat the symptoms of \emph{Shrinkage Bias}. Because they leave the underlying E2M1 grid geometry unchanged, they fail to eliminate its format-level source.

\paragraph*{\heroname{} Recipe.}
Our prior analysis yields a clear design principle: once RHT shifts tensors from a dynamic-range-limited to a local-resolution-limited regime, the 4-bit grid must prioritize local magnitude preservation over extreme dynamic range.
Therefore, we propose \heroname{}, a uniform 4-bit training recipe based on an E1M2/INT4-style uniform grid (\Cref{fig:ufp4-recipe-overview}). 
For every linear-layer GEMM, \heroname{} applies RHT and quantizes the operands to this uniformly spaced grid. 
Crucially, while E2M1 recipes typically restrict RHT to the weight-gradient path to avoid compounding geometric errors, the unbiased nature of \heroname{} allows us to safely enable RHT across all three GEMMs: FPROP (\fwdy{}), DGRAD (\bwddx{}), and WGRAD (\bwddw{}). Stochastic rounding (SR) is applied to $dY$ to preserve gradient expectations. 
Scale hierarchy design remains orthogonal to this format-level solution; one may use single-level, two-level, or block scales depending on hardware efficiency. Our broader recommendation is that future ML accelerators should elevate E1M2/INT4-style uniform grids to first-class training primitives, challenging the status quo that treats E2M1 as the sole viable FP4 format.

\begin{table}[t]
\centering
\small
\caption{
Comparison of E1M2-based recipe \heroname{} with E2M1-based recipe.
}
\label{tab:recipe-config}
\setlength{\tabcolsep}{4pt}
\begin{tabular}{p{0.24\linewidth}p{0.32\linewidth}p{0.32\linewidth}}
\toprule
Configuration & E2M1-based recipe & E1M2-based recipe (\heroname{}) \\
\midrule
Format
& E2M1
& E1M2/INT4-style uniform grid \\
Quant block size
& \(1{\times}16\)
& \(1{\times}16\) \\
Scale hierarchy
& FP32 single-level
& FP32 single-level \\
RHT scope
& \bwddw{}
& \fwdy{}, \bwddx{}, \bwddw{} \\
RHT block size
& 16
& 16 \\
SR scope
& \(dY\)
& \(dY\) \\
2D weight scaling
& \ding{55}
& \ding{55} \\
\bottomrule
\end{tabular}
\end{table}

\paragraph*{Comparison with E2M1-based Recipe.}
\label{sec:compare}

As shown in \Cref{tab:recipe-config}, we keep auxiliary configurations identical across recipes, including block size, scale hierarchy, and SR scope. The methodology differs in only two structural choices: \heroname{} employs a uniform grid to eliminate geometric Shrinkage Bias, which, in turn, permits extending RHT coverage from a single GEMM path (\bwddw{}) to all three (\fwdy{}, \bwddx{}, \bwddw{}). This design directly isolates the impact of the grid format and RHT scope on training stability.

% ======================================================================
\section{Experiments}
\label{sec:experiments}
We organize the experiments as five questions that trace the interaction between FP4 grid geometry and RHT scope from local quantization to end-to-end training:
\begin{itemize}[leftmargin=1.5em]
    \item \textbf{Q1:} Does RHT change the preferred 4-bit grid in both tensor quantization and GEMM outputs?
    \item \textbf{Q2:} Does \heroname{} reduce the BF16-relative training loss gap?
    \item \textbf{Q3:} Does the advantage persist across model scales?
    \item \textbf{Q4:} Which recipe components matter, and can E2M1 approximate a uniform grid?
    \item \textbf{Q5:} Can RHT be fused into FP4 quantization efficiently?
\end{itemize}

\subsection{Experimental setup}
\label{sec:experimental-setup}
% 为了全面并公平的验证我们的猜想，我们首先在 Section 3.1 和 Section 3.2 中对于真实训练过程中采集的 tensor 进行分析，比较在 E2M1/E1M2 格式下，RHT 的引入对于量化误差以及 GEMM 输出结果的影响。
% 我们在 Section 3.3 中进行端到端长跑训练的验证，包括 Dense 1.5B、MoE 7.9B 和 MoE 124B 三个规模的模型。
% 此外，为了确保我们的结论在更大规模下仍然成立，我们在 Section 3.4 中进行了 scaling law 实验，覆盖了 10M-324M 参数规模的 MoE 模型。
% 所有比较在每个实验系列中使用相同的训练数据、优化器、学习率计划、批处理构建、评估节奏和模型配置，以确保结果的公平性和可比性。

% 对于端到端 FP4 训练的实验，我们期望证明 UFP4 所提出的 RHT + E1M2 的设计在长跑下相比于 E2M1 具备持续的优势。
% 在 bf16 精度的基准实验外，我们通过一系列实验找出 E2M1-based recipe 的最佳精度配置，并以此作为 E2M1-based recipe 的参考，而 UFP4 则作为 E1M2-based recipe。
% scale 精度与层级的设计往往和硬件约束相关，它与本文的核心主张正交，因此我们在实验中保持 scale 设计的对齐，对于 E2M1-based recipe 和社区工作（例如 NVFP4 Recipe、xxx）的精度比较可以参考附录 b 中的相关实验。

Within each experimental family, compared runs share data, architecture, optimizer, learning-rate schedule, batch construction, evaluation cadence, and training horizon. For end-to-end FP4 training, we compare the E1M2-based \heroname{} recipe with a strong E2M1 reference selected by a controlled configuration search (\Cref{app:e2m1-search-protocol}); \Cref{tab:recipe-config} summarizes the resulting recipes. We match auxiliary choices such as block size, scale hierarchy, and SR scope whenever possible, so the comparison focuses on the joint effect of FP4 grid choice and RHT scope.

\subsection{Q1: Does RHT change the preferred 4-bit grid in both tensor quantization and GEMM outputs?}
\label{sec:experiments-local}

We first test the interaction between FP4 grid geometry and RHT before end-to-end training. Using MLP tensors collected from real training, we compare E2M1 and E1M2 with and without block RHT at two levels: single-tensor quantization and single-GEMM output.

\begin{figure}[t]
    \centering
    \begin{subfigure}[t]{0.49\linewidth}
        \centering
        \includegraphics[width=\linewidth]{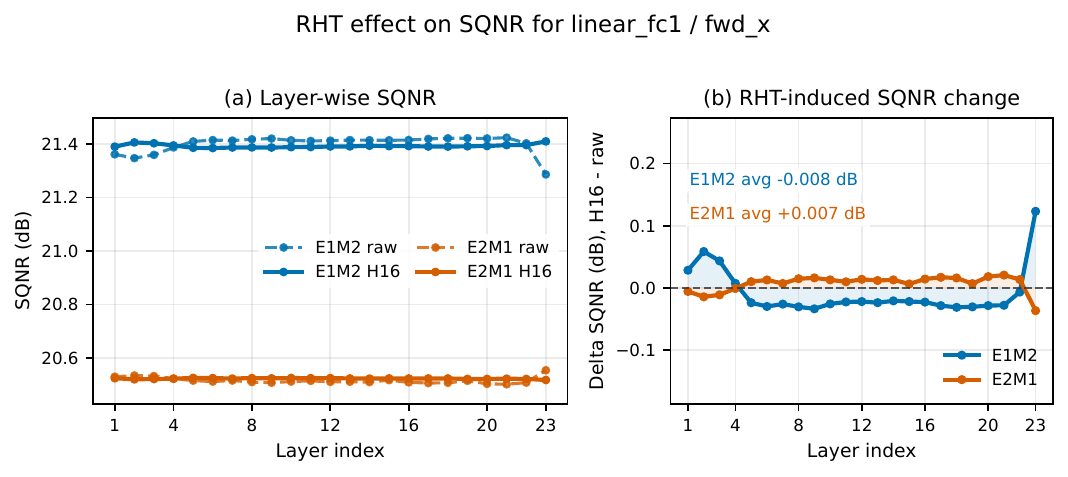}
        \caption{SQNR.}
        \label{fig:single-tensor-fc1-fwdx-sqnr}
    \end{subfigure}
    \hfill
    \begin{subfigure}[t]{0.49\linewidth}
        \centering
        \includegraphics[width=\linewidth]{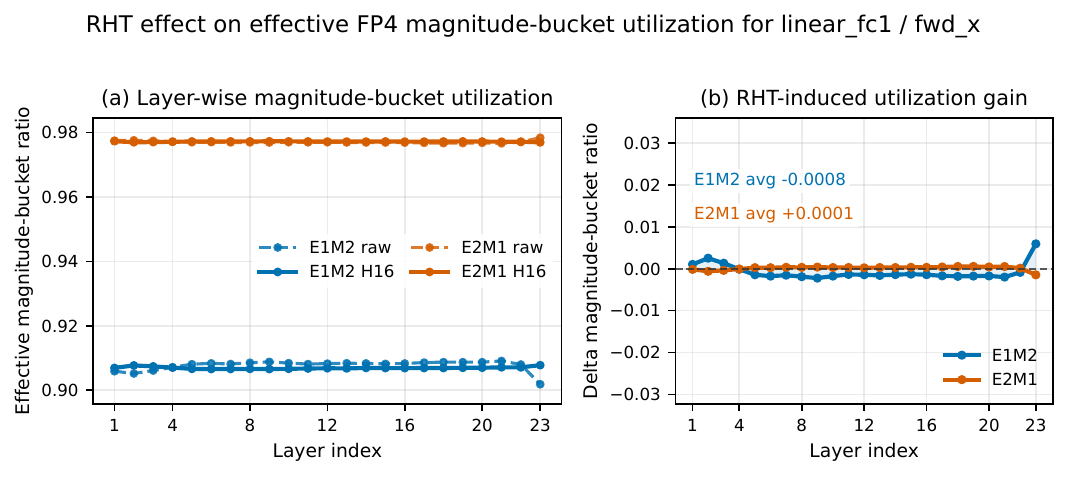}
        \caption{Effective bucket ratio.}
        \label{fig:single-tensor-fc1-fwdx-bucket}
    \end{subfigure}
    \caption{
    Single-tensor quantization diagnostics for the well-behaved \texttt{linear\_fc1}/\texttt{fwd\_x} tensors.
    The left panel of each subfigure reports quantization without RHT and with RHT, while the right panel reports the RHT-induced change.
    }
    \label{fig:single-tensor-fc1-fwdx}
\end{figure}

\begin{figure}[t]
    \centering
    \begin{subfigure}[t]{0.49\linewidth}
        \centering
        \includegraphics[width=\linewidth]{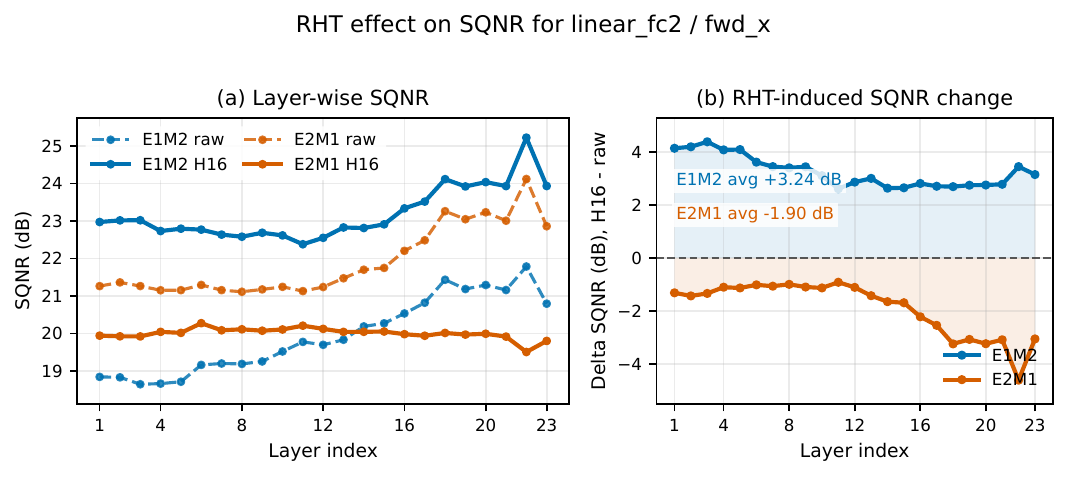}
        \caption{SQNR.}
        \label{fig:single-tensor-fc2-fwdx-sqnr}
    \end{subfigure}
    \hfill
    \begin{subfigure}[t]{0.49\linewidth}
        \centering
        \includegraphics[width=\linewidth]{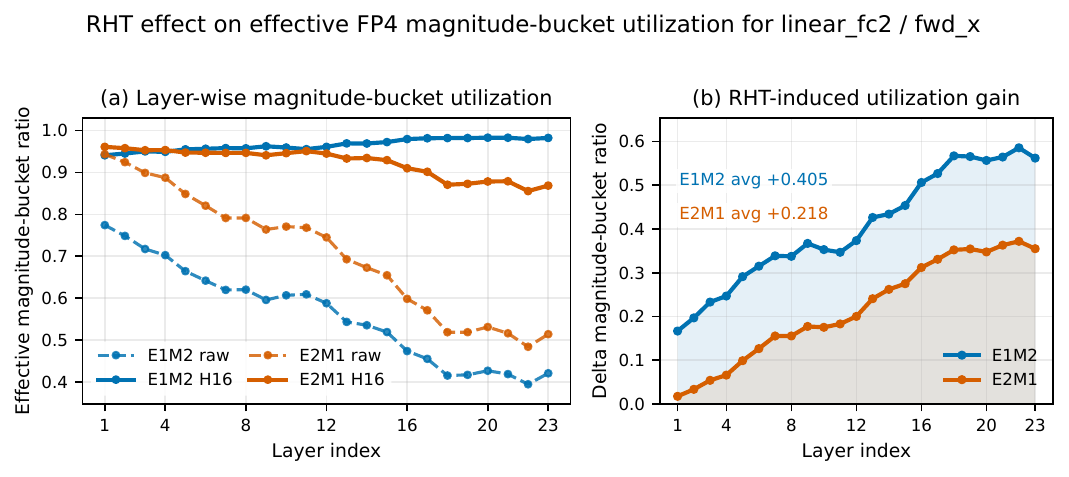}
        \caption{Effective bucket ratio.}
        \label{fig:single-tensor-fc2-fwdx-bucket}
    \end{subfigure}
    \caption{
    Single-tensor quantization diagnostics for the outlier-heavy \texttt{linear\_fc2}/\texttt{fwd\_x} tensors.
    RHT reverses the SQNR ordering: E2M1 is better before rotation, while E1M2 is better after rotation.
    }
    \label{fig:single-tensor-fc2-fwdx}
\end{figure}

\paragraph*{Single-tensor quantization.}
\label{sec:experiments-tensor}
We report SQNR and effective bucket ratio, an entropy-based bucket-utilization metric from \Cref{sec:shrinkage-after-rht}. For the well-behaved \texttt{linear\_fc1}/\texttt{fwd\_x} tensor, RHT is nearly neutral: mean \(\Delta\)SQNR is \(-0.008\) dB for E1M2 and \(+0.007\) dB for E2M1, with negligible effective bucket ratio changes (\(-0.0008\), \(+0.0001\); \Cref{fig:single-tensor-fc1-fwdx}). For outlier-heavy \texttt{linear\_fc2}/\texttt{fwd\_x}, RHT reverses the format ranking: E2M1 leads before rotation (21.90 vs. 19.94 dB), whereas E1M2 leads after rotation (23.19 vs. 20.00 dB) and raises effective bucket ratio from 0.56 to 0.97 on average (\Cref{fig:single-tensor-fc2-fwdx}). Thus, RHT is not universally beneficial; by shifting outlier-heavy tensors from a dynamic-range-limited to a local-resolution-limited regime, it changes the preferred FP4 grid. Extended tensor- and GEMM-level diagnostics across MLP and attention layers are summarized in \Cref{app:rht-sqnr-diagnostics}.

\begin{figure}[H]
    \centering
    \includegraphics[width=0.8\linewidth]{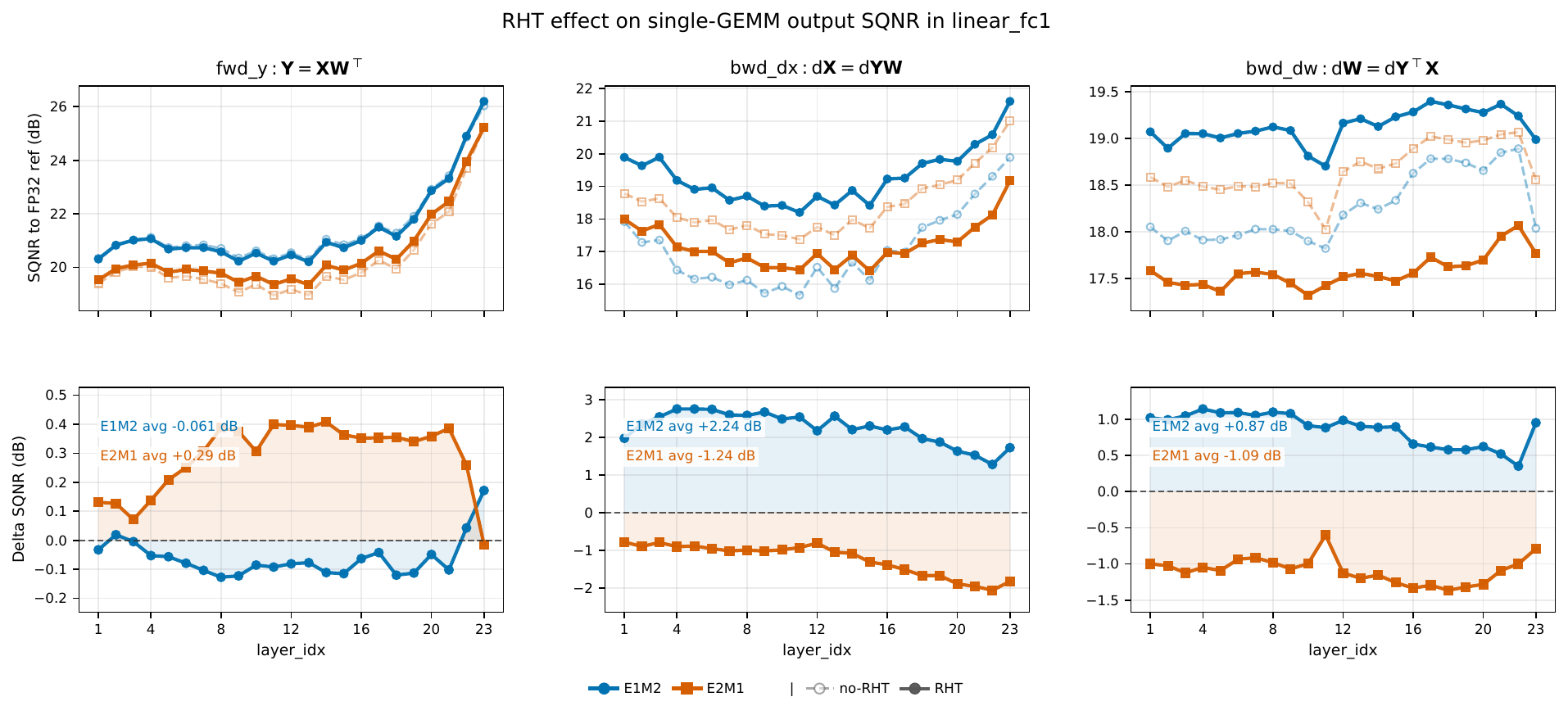}
    \caption{
    Single-GEMM output SQNR for \texttt{linear\_fc1}.
    The bottom row shows the RHT-induced \(\Delta\mathrm{SQNR}\).
    RHT is compatible with E1M2, but can reduce E2M1 output SQNR on the backward GEMMs.
    }
    \label{fig:single-gemm-fc1-output-sqnr}
\end{figure}

\begin{figure}[H]
    \centering
    \includegraphics[width=0.8\linewidth]{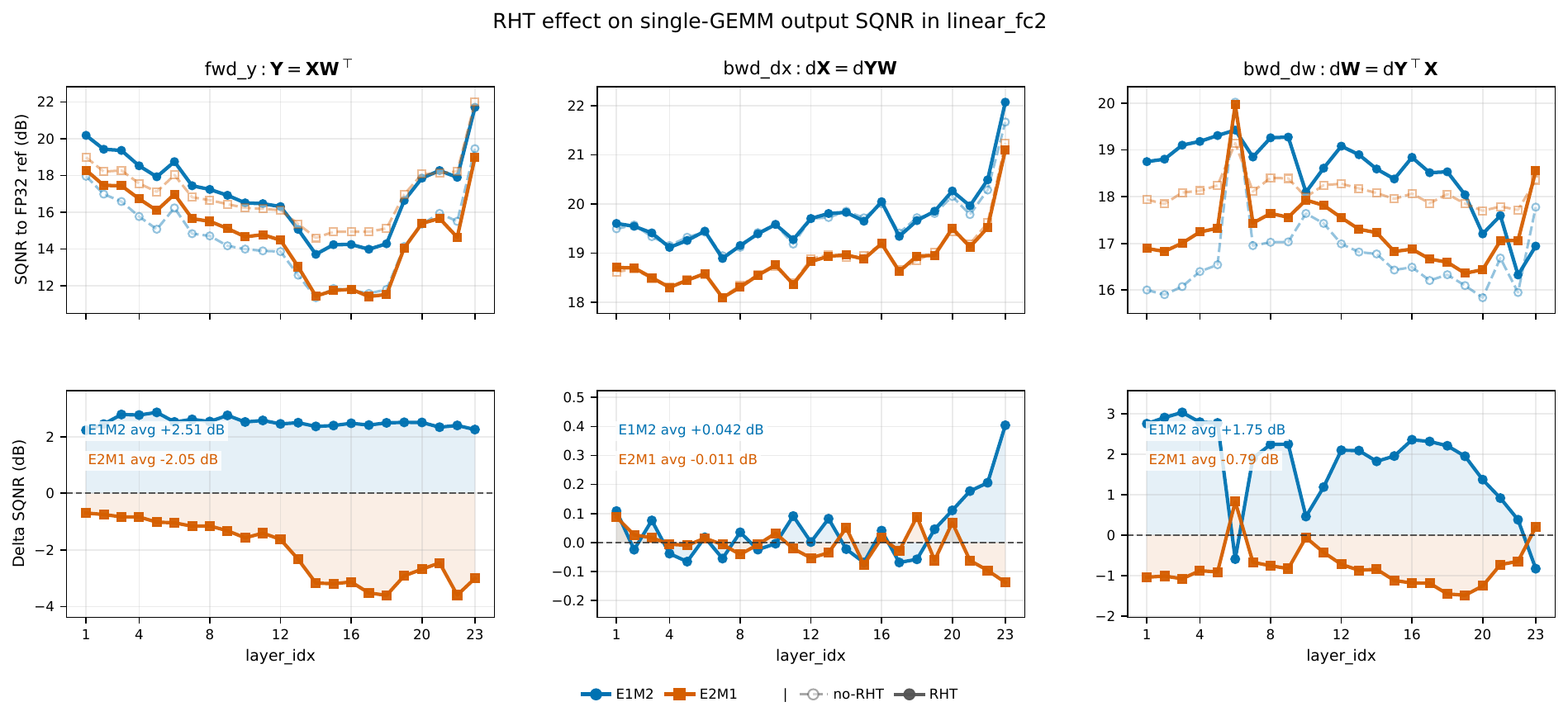}
    \caption{
    Single-GEMM output SQNR for \texttt{linear\_fc2}.
    The format-dependent inversion is more pronounced than in \texttt{linear\_fc1}.
    RHT improves E1M2 output SQNR on the dominant paths, but often decreases E2M1 output SQNR after the operands enter the post-RHT concentrated regime.
    }
    \label{fig:single-gemm-fc2-output-sqnr}
\end{figure}

\paragraph*{Single-GEMM output.}
\label{sec:experiments-gemm}
The same pattern survives GEMM composition. We measure output SQNR for \fwdy{}, \bwddx{}, and \bwddw{} using the same collected tensors. For \texttt{linear\_fc1}, RHT is relatively benign on the forward GEMM but exposes format-dependent degradation on backward paths: E1M2 preserves or improves output SQNR, whereas E2M1 loses SQNR after rotation (\Cref{fig:single-gemm-fc1-output-sqnr}). For \texttt{linear\_fc2}, the inversion is stronger: E1M2 converts post-RHT bucket utilization into higher output SQNR, while E2M1 often degrades after rotation (\Cref{fig:single-gemm-fc2-output-sqnr}). Therefore, the RHT-induced format inversion is not limited to single-tensor diagnostics; it persists in GEMM outputs used by the training computation graph.

\subsection{Q2: Does \heroname{} reduce the BF16-relative training loss gap?}
\label{sec:experiments-longrun}
\begin{figure}[t]
    \centering
    \includegraphics[width=\linewidth]{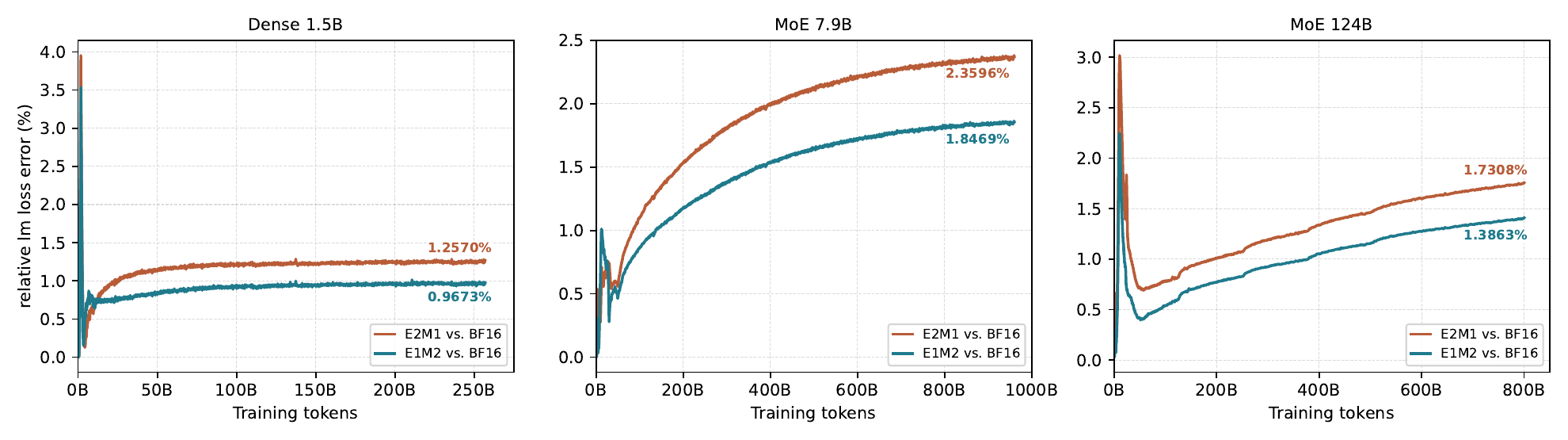}
    \caption{
    The E1M2-based \heroname{} recipe stays closer to BF16 over long-run training.
    Each panel reports the BF16-relative language-modeling loss error
    \(|\mathcal{L}_{r}-\mathcal{L}_{\mathrm{BF16}}|/\mathcal{L}_{\mathrm{BF16}}\)
    for \(r\in\{\mathrm{E2M1},\mathrm{E1M2}\}\).
    Lower is better.
    }
    \label{fig:longrun-relative-loss-error}
\end{figure}

We compare BF16, the E2M1 reference, and the E1M2-based \heroname{} recipe on Dense 1.5B, MoE 7.9B, and MoE 124B long-run pretraining, reporting BF16-relative LM loss error \(|\mathcal{L}_{r}-\mathcal{L}_{\mathrm{BF16}}|/\mathcal{L}_{\mathrm{BF16}}\). Across all three settings, \heroname{} stays closer to BF16 (\Cref{fig:longrun-relative-loss-error}): latest-1000-step relative error drops from 1.2570\% to 0.9673\% on Dense 1.5B, from 2.3596\% to 1.8469\% on MoE 7.9B, and from 1.7308\% to 1.3863\% on MoE 124B. Thus, the local interaction between FP4 grid geometry and RHT scope observed in Q1 persists in long-run training, although both FP4 recipes still incur a measurable BF16 gap.

% \begin{table}[t]
% \centering
% \small
% \caption{
% Endpoint BF16-relative language-modeling loss error.
% Relative errors are computed against the matched BF16 run and averaged over the latest 1000 matched steps for each model family.
% Lower is better.
% }
% \label{tab:longrun-relative-loss-error}
% \begin{tabular}{lccc}
% \toprule
% Model & E2M1 relative error & E1M2/\heroname{} relative error & Abs. reduction \\
% \midrule
% Dense 1.5B & 1.2570\% & 0.9673\% & 0.2897\% \\
% MoE 7.9B & 2.3596\% & 1.8469\% & 0.5127\% \\
% MoE 124B & 1.7308\% & 1.3863\% & 0.3445\% \\
% \bottomrule
% \end{tabular}
% \end{table}

\subsection{Q3: Does the advantage persist across model scales?}
\label{sec:experiments-scaling}
\begin{figure}[t]
    \centering
    \begin{subfigure}[t]{0.5\linewidth}
        \centering
        \includegraphics[width=\linewidth]{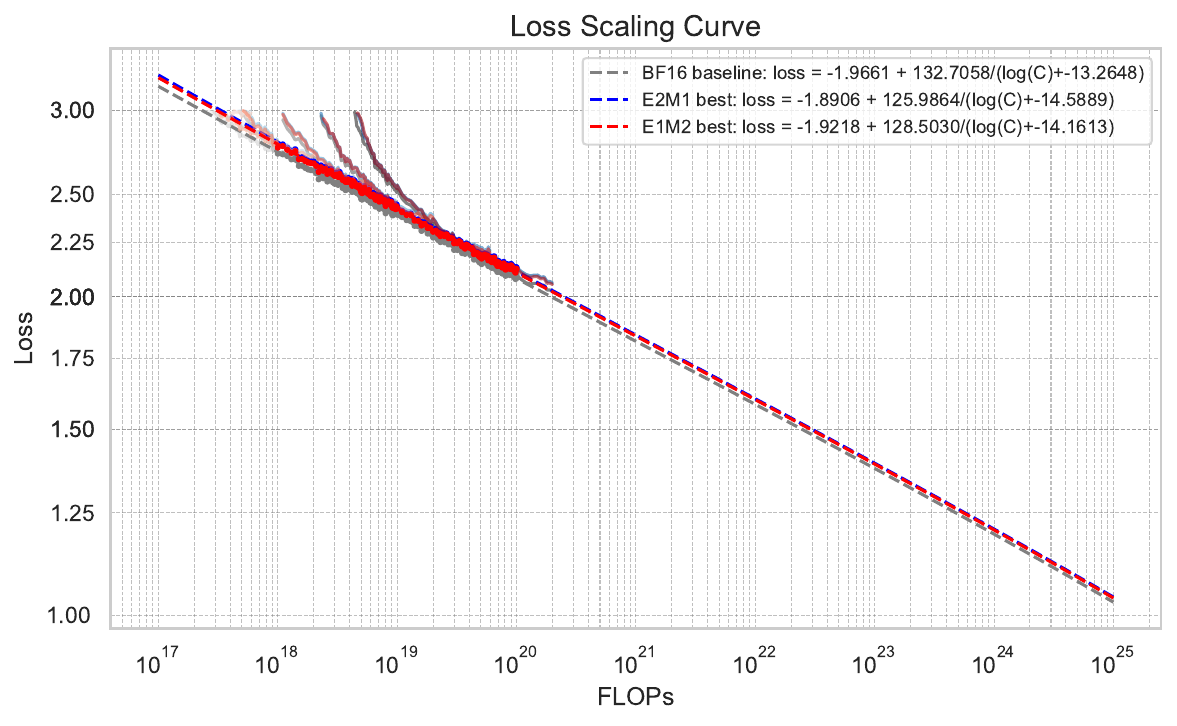}
        \caption{Scaling-law fit and extrapolation.}
        \label{fig:scaling-laws-analysis-curve}
    \end{subfigure}
    \hfill
    \begin{subfigure}[t]{0.4\linewidth}
        \centering
        \includegraphics[width=\linewidth]{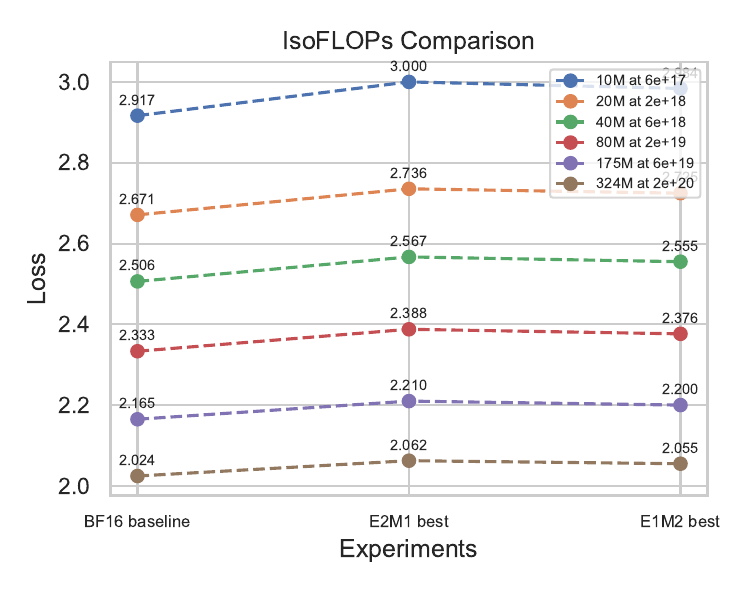}
        \caption{Loss comparison across model sizes.}
        \label{fig:scaling-laws-analysis-loss-compare}
    \end{subfigure}
    \caption{
    Scaling-law validation of the E1M2 recipe (\heroname{}) against the E2M1 reference.
    The fitted curve summarizes the trend across the scaling sweep, while the loss comparison shows the observed gap at the measured model sizes.
    }
    \label{fig:scaling-laws-analysis}
\end{figure}

Following the Ling scaling-law protocol~\citep{Tian2025TowardsGL}, we train 10M--324M MoE models under matched data, optimization, and compute budgets, and fit loss as a function of training compute. Across the measured range and fitted extrapolation, the E1M2 curve remains below the E2M1 reference (\Cref{fig:scaling-laws-analysis}), indicating that the advantage is not confined to the smallest models. The fitted FP4-to-BF16 gap also decreases with compute, suggesting that the FP4 penalty does not grow over the scaling sweep, although a residual gap to BF16 remains.

\subsection{Q4: Which recipe components matter, and can E2M1 approximate a uniform grid?}

\paragraph*{RHT scope and stochastic rounding.}
\label{sec:experiments-ablations}
We ablate RHT scope and SR on Dense 1.5B E1M2 FP4 runs trained beyond 100B tokens (\Cref{tab:e1m2-rht-sr-ablation}). Full-RHT coverage is best: relative to no RHT, \bwddw{}-only RHT reduces loss by 0.00481, \fwdy{}+\bwddw{} by 0.00644, \bwddx{}+\bwddw{} by 0.00290, and full RHT by 0.01123. With full RHT fixed, SR on \(dY\) gives another 0.00456 reduction. Thus, both full RHT and SR contribute. This result is notable because prior NVFP4 recipes typically avoid rotating non-leaf GEMM outputs such as \fwdy{} and \bwddx{}, where RHT can be harmful under E2M1-style grids \citep{nvidia2026pretraininglargelanguagemodels,chen2026tetrajetv2accuratenvfp4training}; our ablation supports the mechanism: once the post-RHT regime is represented by a uniform grid, full-RHT coverage becomes beneficial in this setting.

\begin{table}[t]
\centering
\caption{
Ablation study of RHT scope and SR under FP4 E1M2 training.
For the RHT ablation block, $\Delta$ is measured against SR enabled with no RHT.
For the SR ablation block, $\Delta$ is measured against the full-RHT, no-SR variant.
}
\label{tab:e1m2-rht-sr-ablation}
\resizebox{0.8\linewidth}{!}{
\begin{tabular}{lcccccc}
\toprule
Setting & SR & \texttt{fwd\_y} & \texttt{bwd\_dx} & \texttt{bwd\_dw}
& Mean LM loss & $\Delta$ loss \\
\midrule
\multicolumn{7}{l}{\textit{RHT scope ablation}} \\
No RHT
& \checkmark & -- & -- & --
& 1.89202 & 0.00000 \\
RHT on \texttt{bwd\_dw}
& \checkmark & -- & -- & \checkmark
& 1.88721 & -0.00481 \\
RHT on \texttt{bwd\_dx}, \texttt{bwd\_dw}
& \checkmark & -- & \checkmark & \checkmark
& 1.88912 & -0.00290 \\
RHT on \texttt{fwd\_y}, \texttt{bwd\_dw}
& \checkmark & \checkmark & -- & \checkmark
& 1.88558 & -0.00644 \\
\textbf{Full RHT w/ SR (\heroname{})}
& \checkmark & \checkmark & \checkmark & \checkmark
& \textbf{1.88079} & \textbf{-0.01123} \\
\midrule
\multicolumn{7}{l}{\textit{SR ablation under full RHT}} \\
Full RHT w/o SR
& -- & \checkmark & \checkmark & \checkmark
& 1.88535 & 0.00000 \\
\textbf{Full RHT w/ SR (\heroname{})}
& \checkmark & \checkmark & \checkmark & \checkmark
& \textbf{1.88079} & \textbf{-0.00456} \\
\bottomrule
\end{tabular}
}
\end{table}

\begin{figure}[t]
    \centering
    \includegraphics[width=0.8\linewidth]{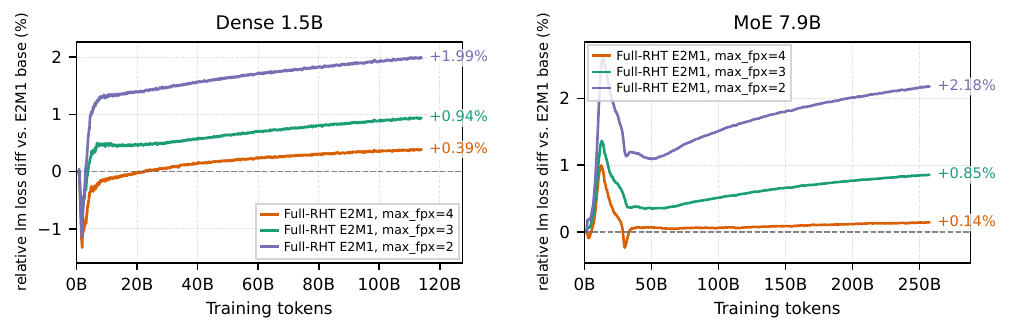}
    \caption{
    Relative LM loss of full-RHT E2M1 variants compared to the E2M1 reference.
    }
    \label{fig:e2m1-max-fpx-ablation}
\end{figure}

\paragraph*{Can range-restricted E2M1 emulate a uniform grid?}
Current FP4 hardware paths may be tied to E2M1/NVFP4-style data elements, so we test whether range-restricted E2M1 can emulate uniform-grid behavior. The E2M1 reference uses \(\texttt{max\_fpx}=6\) and RHT only on \bwddw{}; full-RHT variants reduce \(\texttt{max\_fpx}\), with \(\texttt{max\_fpx}=2.0\) retaining only \(\{0,0.5,1.0,1.5,2.0\}\). Although this avoids high-magnitude asymmetric bins, it also sacrifices dynamic range and bucket utilization: all tested range-restricted variants underperform the E2M1 reference on Dense 1.5B and MoE 7.9B (\Cref{fig:e2m1-max-fpx-ablation}). In these tested recipes, E2M1 range restriction is therefore not a satisfactory substitute for native E1M2/INT4 support.

Recent HiFloat4 work~\citep{taghian2026hifloat4formatlanguagemodel} adopts a uniform S1P2 data element, making future Ascend 960 systems~\citep{huaweiGroundbreakingSuperPoD} a promising platform for realizing \heroname{} natively. Our recommendation is narrow: E2M1 should remain available for raw outlier-heavy tensors and inference workloads, but training hardware should expose E1M2/INT4-style uniform grids as first-class FP4 training data elements.

\subsection{Q5: Can RHT be fused into FP4 quantization efficiently?}
\label{sec:experiments-performance}
Full RHT adds only a small block Hadamard transform before FP4 quantization. When the Hadamard block matches the quantization block, the transform can be fused before scale estimation and packing, avoiding an intermediate rotated tensor. For block size 16, the fused RHT +quantization is about \(1.06\times\) and \(1.07\times\) the latency of standalone quantization across tested BF16 matrix shapes on SM90 and SM100, respectively, whereas unfused RHT+quantization has \(1.62\times\) and \(1.41\times\) the fused latency. Thus, full-RHT quantization has small fused-kernel overhead, although end-to-end training overhead still depends on full-system integration.

\begin{figure}[t]
    \centering
    \includegraphics[width=0.8\linewidth]{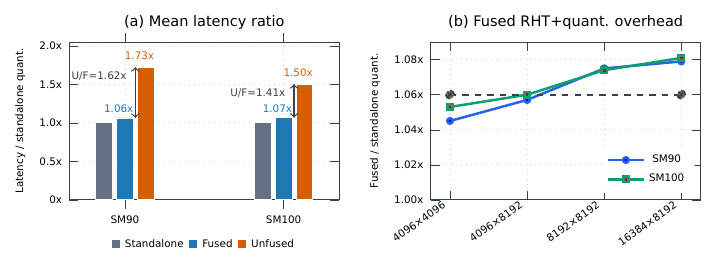}
    \caption{
    Relative latency of RHT+quantization.
    Left: mean latency ratio relative to standalone quantization; bar heights and labels are normalized by standalone quantization, while gray arrows report U/F (unfused-to-fused) ratios.
    Right: per-shape fused RHT+quantization overhead.
    }
    \label{fig:f8-fused-rht-quant-throughput-v0}
\end{figure}

\section{Related Work}
\label{sec:related-work}

\paragraph*{FP4 formats, block scaling, and scale hierarchy.}
A first line of FP4 training work improves quantization quality by refining the data format, block size, and scale hierarchy.
MXFP4 and NVFP4 use fine-grained block scaling, and NVFP4 further uses smaller blocks and a two-level scale hierarchy to improve E2M1 training accuracy~\citep{rouhani2023microscaling,nvidia2026pretraininglargelanguagemodels}.
Recent studies show that this design is not limited to a fixed E2M1 path: HiFloat4 adopts a uniform S1P2 data element with hierarchical scales on Ascend NPUs~\citep{taghian2026hifloat4formatlanguagemodel}, MixFP4 adaptively chooses E2M1 or E1M2 per block within an NVFP4-compatible scale hierarchy~\citep{zou2026mixfp4}, and Four Over Six improves NVFP4 by evaluating alternative block scales to better represent near-maximal values~\citep{cook2026sixaccuratenvfp4quantization}.
INT-vs-FP comparisons further show that fine-grained INT formats can become competitive with FP formats once block size and outlier mitigation are considered~\citep{chen2025intvsfpcomprehensive}.
Together, these works show that scale hierarchy, block size, and data format design are all critical for practical FP4 training.
\heroname{} studies a complementary axis: under matched block size and scale hierarchy, it isolates whether post-RHT tensors should be quantized to non-uniform E2M1 or to an E1M2/INT4-style uniform grid.

\paragraph*{Quantizer-side and training-aware methods.}
A second line modifies the quantizer or training estimator to make FP4 training more stable.
Microsoft FP4 introduces a differentiable quantization estimator and an outlier clamping/compensation strategy to prevent activation collapse~\citep{2025arXiv250117116W};
Quartet II improves unbiased gradient estimation with a microscaling EDEN routine~\citep{panferov2026quartetiiaccuratellm};
FAAR learns format-aware adaptive rounding decisions by explicitly accounting for the non-uniform E2M1 grid~\citep{li2026faarformatawareadaptiverounding};
and TetraJet-v2 adds backward alignment, stochastic-rounding changes, oscillation suppression, and outlier control to NVFP4 training~\citep{chen2026tetrajetv2accuratenvfp4training}.
These methods reduce estimator bias, rounding error, variance, or optimization instability, and are complementary to \heroname{}: improved estimators, adaptive rounding rules, or stabilizers can be combined with a uniform grid.

\paragraph*{Tensor-side preprocessing.}
Another line changes tensors before quantization through mathematically equivalent rewrites.
Rotation-based methods such as RHT, QuaRot, SpinQuant, and FlatQuant disperse outlier energy before quantization~\citep{3737916.3741096,liu2025spinquantllmquantizationlearned,sun2025flatquant}.
Tensor-decomposition methods reduce low-bit error by decomposing tensors before quantization; related approaches include mean--SVD-style decompositions, outlier-channel separation, and smoothing methods~\citep{cao2025metis,xiao2023smoothquant,lisvdquant,chen2026tetrajetv2accuratenvfp4training}.
These methods act upstream of the quantizer: they change the distribution presented to the format grid.
In this paper, we show that a post-RHT tensor may still suffer from \emph{Shrinkage Bias} when quantized to non-uniform E2M1.
This limitation is imposed by the format grid itself.
Therefore, under E1M2/INT4-style uniform formats, the same tensor-side preprocessing methods may be better able to translate improved tensor distributions into quantization-quality gains.
% 在本文中，我们验证了 a post-RHT tensor 在量化到 E2M1 时会碰到 non-uniform bin shrinkage bias。
% 这是 format grid 本身的约束，在 E1M2/INT4-Style format 下，类似的 tensor preprocessing 有潜力发挥更好的效果。

\section{Conclusion}
\label{sec:conclusion}

We revisit the default use of E2M1 for FP4 training under RHT-based outlier mitigation. 
Our analysis shows that RHT changes the quantization regime from dynamic-range-limited to local-resolution-limited, where the asymmetric RTNE bins of E2M1 can introduce Shrinkage Bias. 
This effect is visible in real tensor and GEMM diagnostics, and the same format-dependent pattern persists in long-run dense and MoE training. 
By using an E1M2/INT4-style uniform grid, applying RHT to all three training GEMMs, and using stochastic rounding only on \(dY\), \heroname{} consistently reduces the BF16-relative FP4 loss error compared with the E2M1 reference recipe.

Our implication is narrow but actionable: E2M1 should remain available for range-limited workloads, but it should not be the only first-class FP4 training format. 
Future training accelerators should support E1M2/INT4-style uniform 4-bit grids as first-class training data elements, enabling recipes such as \heroname{} to combine post-RHT numerical stability with native 4-bit matrix throughput.

% References.
\bibliographystyle{assets/plainnat}
\bibliography{main}

\appendix
\section*{Appendix}
\section{RHT-Induced SQNR Changes in Tensors and GEMM Outputs}
\label{app:rht-sqnr-diagnostics}

The main text uses layerwise MLP traces to explain how RHT changes the preferred FP4 grid.  Here we aggregate the same \(\Delta\mathrm{SQNR}\) diagnostic over MLP and attention layer families, covering both single-tensor quantization and single-GEMM outputs.

\begin{figure*}[h]
    \centering
    \begin{subfigure}[t]{0.495\linewidth}
        \centering
        \includegraphics[width=\linewidth]{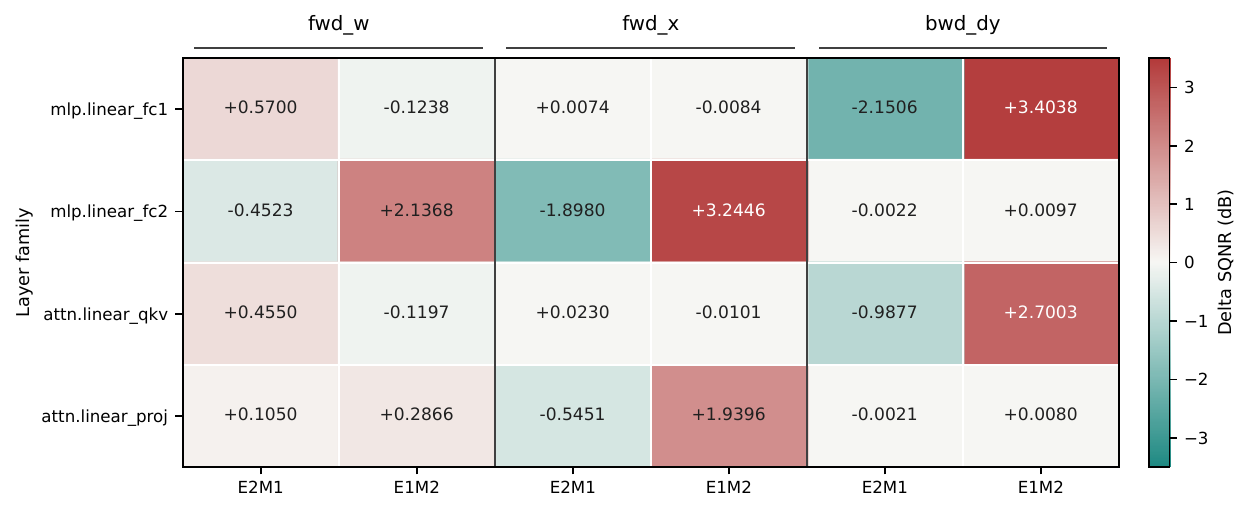}
        \caption{Single-tensor quantization.}
        \label{fig:app-single-tensor-sqnr-delta}
    \end{subfigure}\hfill
    \begin{subfigure}[t]{0.495\linewidth}
        \centering
        \includegraphics[width=\linewidth]{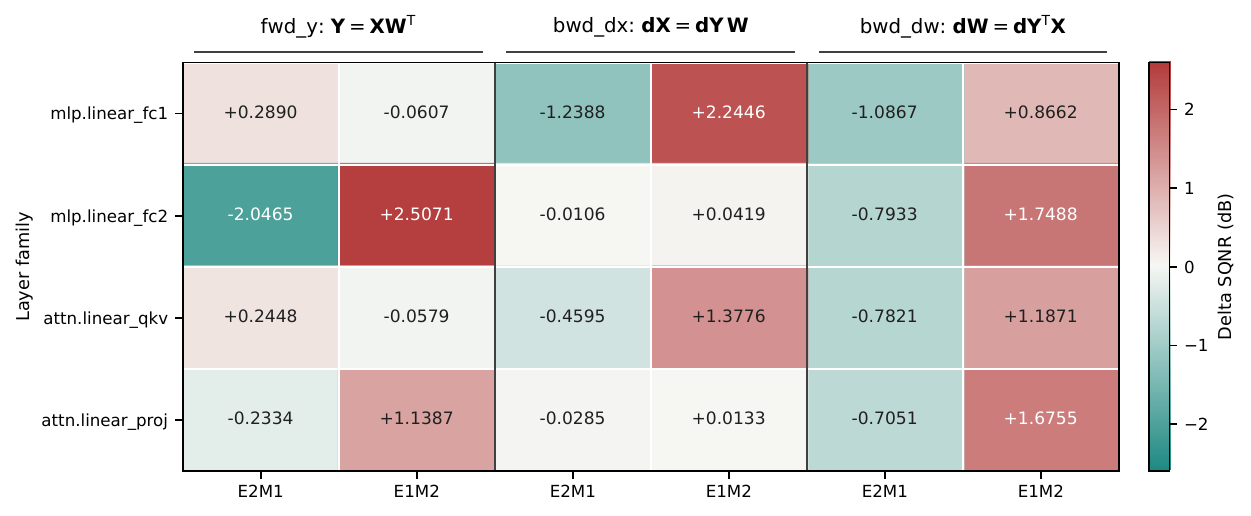}
        \caption{Single-GEMM output.}
        \label{fig:app-single-gemm-sqnr-delta}
    \end{subfigure}
    \caption{
    Mean RHT-induced \(\Delta\mathrm{SQNR}\) for tensor quantization and GEMM outputs.
    Rows average over 23 layers; positive/negative values indicate SQNR gain/loss after RHT.
    }
    \label{fig:app-rht-sqnr-delta-summary}
\end{figure*}

\Cref{fig:app-rht-sqnr-delta-summary} gives two takeaways.  First, the RHT effect is format-dependent: E1M2 produces large gains on rotation-sensitive tensors (e.g., \texttt{linear\_fc2}/\texttt{fwd\_x}) and remains near-neutral on already well-behaved ones (e.g., \texttt{linear\_fc1}/\texttt{fwd\_x}), whereas E2M1 often loses SQNR on these rotation-sensitive tensors.  This is the empirical signature expected from the shrinkage-bias mechanism: after RHT, local resolution matters more than excess dynamic range.

Second, the same format-ranking inversion appears in several GEMM outputs and extends to attention modules.  This is not a blanket failure of RHT under E2M1: some E2M1 FPROP outputs show small SQNR gains, such as \texttt{linear\_fc1} and \texttt{linear\_qkv}.  Rather, E2M1 degradation is concentrated on rotation-sensitive paths, including \texttt{linear\_fc2} FPROP (\fwdy{}) and attention backward GEMMs, whereas E1M2 yields positive \(\Delta\mathrm{SQNR}\) on the main rotation-sensitive paths, supporting full-RHT coverage in \heroname{}.

\section{Controlled E2M1 Reference Selection}
\label{app:e2m1-search-protocol}

To avoid an undertuned E2M1 baseline, we tune its training configuration through controlled one-factor ablations over the dimensions in \Cref{tab:e2m1-search-space}.  Each candidate run trains Dense 1.5B for 200B tokens with identical data, architecture, optimizer, learning-rate schedule, and all non-ablated recipe settings.  Within each ablation, we choose the setting with the lowest average LM loss over the final 1,000-step window.  The resulting E2M1 configuration is then frozen for the long-run, scaling, and component-ablation experiments.

The selected E2M1 structure is WGRAD-only RHT (\bwddw{}) with \(dY\)-only SR.  Under E2M1, FPROP RHT (\fwdy{}) remains harmful, while DGRAD RHT (\bwddx{}) is initially neutral but yields higher LM loss at later training stages; the selected RHT scope is therefore \bwddw{}.  For stochastic rounding, \(dY\)-only SR achieves the lowest LM loss.  Adding SR to the backward GEMM operands \(X_{\mathrm{WGRAD}}\) and \(W_{\mathrm{DGRAD}}\) slightly increases LM loss in our E2M1 setting.  This observation differs from the NVFP4 setting studied by \tetrajet{}~\citep{chen2026tetrajetv2accuratenvfp4training}, where applying SR beyond \(dY\) to additional backward GEMM operands is reported to improve training.  We therefore use the best observed E2M1 setting from these ablations as the reference configuration.

\begin{table}[t]
\centering
\small
\caption{
Controlled one-factor ablation space for selecting the E2M1 reference.
For a set \(\mathcal S\), \(2^{\mathcal S}\) denotes its power set, i.e., the set of all combinations formed from elements of \(\mathcal S\), including the empty combination.
}
\label{tab:e2m1-search-space}
\resizebox{\linewidth}{!}{
\begin{tabular}{
>{\raggedright\arraybackslash}m{0.20\linewidth}
>{\raggedright\arraybackslash}m{0.30\linewidth}
>{\raggedright\arraybackslash}m{0.15\linewidth}
>{\raggedright\arraybackslash}m{0.45\linewidth}}
\toprule
Dimension & Search space & Selected settings & Notes \\
\midrule
RHT scope
& $2^{\{\fwdy{},\bwddx{},\bwddw{}\}}$
& \texttt{bwd\_dw}
& WGRAD-only selected; FPROP RHT is detrimental and DGRAD RHT raises LM loss at later stages. \\
RHT block size
& 16 / 32 / 64 / 128
& 16
& Comparable LM loss; 16 aligns with quant blocks and facilitates fused RHT+quantization. \\
SR scope
& $2^{\{ dY, X_{\mathrm{FPROP}}, W_{\mathrm{FPROP}}, X_{\mathrm{WGRAD}}, W_{\mathrm{DGRAD}} \}}$
& \(dY\)
& \(dY\)-only selected; backward \(X_{\mathrm{WGRAD}}, W_{\mathrm{DGRAD}}\) SR slightly increases LM loss. \\
2D weight scaling
& \ding{51} / \ding{55}
& \ding{55}
& Disabled. \\
\texttt{max\_fpx}
& 4 / 6
& 6
& Maximum FP4 magnitude; 6 preserves the full E2M1 range and minimizes LM loss. \\
\bottomrule
\end{tabular}
}
\end{table}

\section{Exponential Approximation of Multiplicative Accumulation}
\label{sec:appendix-propagation}

Here we justify the exponential approximation used in \Cref{eq:multi-gemm-attenuation}.
Let \(\delta_k=1-\eta_k\) denote the per-GEMM multiplicative loss, so that the cumulative multiplicative factor is
\begin{equation}
    \prod_{k=1}^{K}\eta_k
    =
    \prod_{k=1}^{K}(1-\delta_k).
    \label{eq:app-product-form}
\end{equation}
Taking the logarithm of the product gives
\begin{equation}
    \begin{aligned}
    \prod_{k=1}^{K}(1-\delta_k)
    &=
    \exp\!\left(\log\prod_{k=1}^{K}(1-\delta_k)\right)
    \\
    &=
    \exp\!\left(\sum_{k=1}^{K}\log(1-\delta_k)\right) \\
    &\approx
    \exp\!\left(
        -\sum_{k=1}^{K}\delta_k
    \right).
    \end{aligned}
    \label{eq:app-exponential-accumulation}
\end{equation}
For small per-GEMM multiplicative loss, the logarithm admits the Taylor expansion
\begin{equation}
    \log(1-\delta_k)
    =
    -\delta_k
    +
    O\!\left(\delta_k^2\right),
    \qquad |\delta_k|\ll1 .
    \label{eq:app-log-taylor}
\end{equation}
Substituting this expansion into \Cref{eq:app-exponential-accumulation} yields
\begin{equation}
    \prod_{k=1}^{K}(1-\delta_k)
    =
    \exp\!\left(
    -\sum_{k=1}^{K}\delta_k
    +
    O\!\left(\sum_{k=1}^{K}\delta_k^2\right)
    \right).
    \label{eq:app-exponential-remainder}
\end{equation}
When the second-order term \(O(\sum_k\delta_k^2)\) is small relative to the first-order accumulation, we obtain the approximation
\begin{equation}
    \prod_{k=1}^{K}(1-\delta_k)
    \approx
    \exp\!\left(
    -\sum_{k=1}^{K}\delta_k
    \right).
    \label{eq:app-exponential-first-order}
\end{equation}
This shows that even small per-GEMM multiplicative losses add in the exponent. Consequently, a weak but consistently positive \(\delta_k\) can produce a visible cumulative decay over long computation paths.

\end{document}